\documentclass[authoryear,preprint,12pt]{elsarticle}
\usepackage{amssymb}
\usepackage{natbib}
\usepackage{xcolor}
\usepackage{subcaption}  % 或者 \usepackage{subfigure}
\usepackage{tcolorbox}  
\usepackage{amsmath}

\definecolor{darkred}{RGB}{200, 0, 0}
\definecolor{darkyellow}{RGB}{250, 153, 0}
\definecolor{darkgreen}{RGB}{0, 180, 0}
\definecolor{lise}{RGB}{128,0,32}
\definecolor{deepblue}{RGB}{68,85,102}
\definecolor{graygreen}{RGB}{85,107,47}
\definecolor{tomato}{RGB}{255,99,71}
\definecolor{india}{RGB}{205,92,92}

\usepackage{hyperref}
\hypersetup{
    colorlinks=true,       % false: boxed links; true: colored links
    linkcolor=RoyalBlue,   % color of internal links
    citecolor=[RGB]{128,0,32},  % 深栗色的引用
    filecolor=Mahogany,    % color of file links
    urlcolor=Cerulean      % color of external urls
}

\journal{Nuclear Physics B}

\begin{document}

\begin{frontmatter}



\title{Beyond Frequency: The Role of Redundancy in Large Language Model Memorization}

\author[inst1]{Jie Zhang\corref{cor1}}
\ead{jiezhang.spark@gmail.com}
\cortext[cor1]{Corresponding author}

\author[inst4]{Qinghua Zhao}
\ead{lshowway@gmail.com}

\author[inst1]{Chi-ho Lin}
\ead{ich410@semyung.ac.kr}

\author[landa]{Zhongfeng Kang}
\ead{kangzf@lzu.edu.cn}

\author[inst2,inst3]{Lei Li}
\ead{lilei@di.ku.dk}

\affiliation[inst1]{organization={Semyung University},%Department and Organization
            % addressline={Address One}, 
            % city={City One},
            % postcode={00000}, 
            % state={State One},
            country={Korea},
            }
% , Republic of
\affiliation[inst2]{organization={University of Copenhagen},%Department and Organization
            % addressline={Address Two}, 
            city={Copenhagen},
            % postcode={22222}, 
            % state={State Two},
            country={Denmark}
            }
\affiliation[inst3]{organization={University of Washington},%Department and Organization
            % addressline={Address Two}, 
            city={Seattle},
            % postcode={22222}, 
            % state={State Two},
            country={USA}
            }
\affiliation[inst4]{organization={SAIBD, Hefei University},%Department and Organization
            % addressline={Address Two}, 
            city={Hefei},
            % postcode={22222}, 
            % state={State Two},
            country={China}
            }
\affiliation[landa]{organization={School of Information Science and Engineering, Lanzhou University},%Department and Organization
            % addressline={Address Two}, 
            city={Lanzhou},
            postcode={730000}, 
            % state={State Two},
            country={China}
            }

\begin{abstract}
Memorization in large language models poses critical risks for privacy and fairness as these systems scale to billions of parameters.  While previous studies established correlations between memorization and factors like token frequency and repetition patterns, we revealed distinct response patterns: frequency increases minimally impact memorized samples (e.g. 0.09) while substantially affecting non-memorized samples (e.g., 0.25), with consistency observed across model scales. Through  counterfactual analysis by perturbing sample prefixes and quantifying perturbation strength through token positional changes, we demonstrate that  redundancy correlates with memorization patterns. Our findings establish that: $\sim$79\% of memorized samples are low-redundancy, these low-redundancy samples exhibit 2-fold higher vulnerability than high-redundancy ones, and consequently memorized samples drop by 0.6 under perturbation while non-memorized samples drop by only 0.01, indicating that more redundant content becomes both more memorable and more fragile. These findings suggest potential redundancy-guided approaches for data preprocessing, thereby reducing privacy risks and mitigating bias to ensure fairness in model deployments.
\end{abstract}


\end{frontmatter}

\section{Introduction}

Large language models (LLMs) can verbatim memorize training sequences, reproducing exact content from their training corpora. 
The memorization capabilities of LLMs have emerged as a double-edged sword in AI development. While enabling impressive performance on knowledge-intensive tasks, these same capabilities create substantial risks for privacy leakage, bias amplification, and adversarial vulnerabilities \citep{carlini2021extracting, karamolegkou2023copyright, LEHMLER2024128473, luo2025sharedpathunravelingmemorization}. Understanding and controlling this memorization behavior has become critical as models scale to trillions of parameters and process increasingly sensitive data.
Research on memorization mechanisms enables practical improvements in model training: corpus deduplication reduces memorization while maintaining performance \citep{lee-etal-2022-deduplicating, sakarvadia2025mitigating}, prevents privacy risks \citep{kandpal2022deduplicating, kassem-etal-2023-preserving, staab2024beyond}, and mitigates bias amplification \citep{10.1007/978-981-96-0567-5_24}.

Memorization research spans multiple investigative directions. At the mechanistic level, studies have  identified MLP neurons for factual storage \citep{geva-etal-2023-dissecting, bayazit-etal-2024-discovering, zhu-etal-2024-fastmem} and layer-specific functions \citep{haviv-etal-2023-understanding}, with recent work revealing spatial differentiation of memorization mechanisms \citep{huang2025neuronleveldifferentiationmemorizationgeneralization}.
Meanwhile, content analyses have revealed preferential memorization of nouns, numbers, and special characters \citep{bai2024specialcharactersattackscalable}, while scaling studies established  relationships between memorization and model size, training duration, and data frequency \citep{tirumala2022memorization, carlini2023quantifying, lu-etal-2024-scaling, 10.5555/3692070.3693506}. However, existing work has primarily focused on surface-level characteristics, leaving more implicit factors affecting memorization behavior unanswered, such as information redundancy, i.e., a fundamental property determining content resilience to noise.

We discovered a critical disparity  that challenges conventional understanding: established factors (e.g., frequency, repetition) exhibit different effects on memorized versus non-memorized samples. Increases in these factors minimally impact memorized samples' memorization scores but substantially affect non-memorized samples. For instance, as token frequency increases from 0 to 8, in Pythia 12B, memorized samples' scores decrease by 0.09 while non-memorized samples' scores increase by 0.25. 
% Similarly, as repetition count increases from 0 to 60, memorized samples' scores increase by 0.04 while non-memorized samples' scores decrease by 0.16. 
This disparity suggests that surface-level characteristics are symptoms and deep-level characteristics are needed.

 \begin{figure}[!t]
  \centering
  \includegraphics[width=\linewidth]{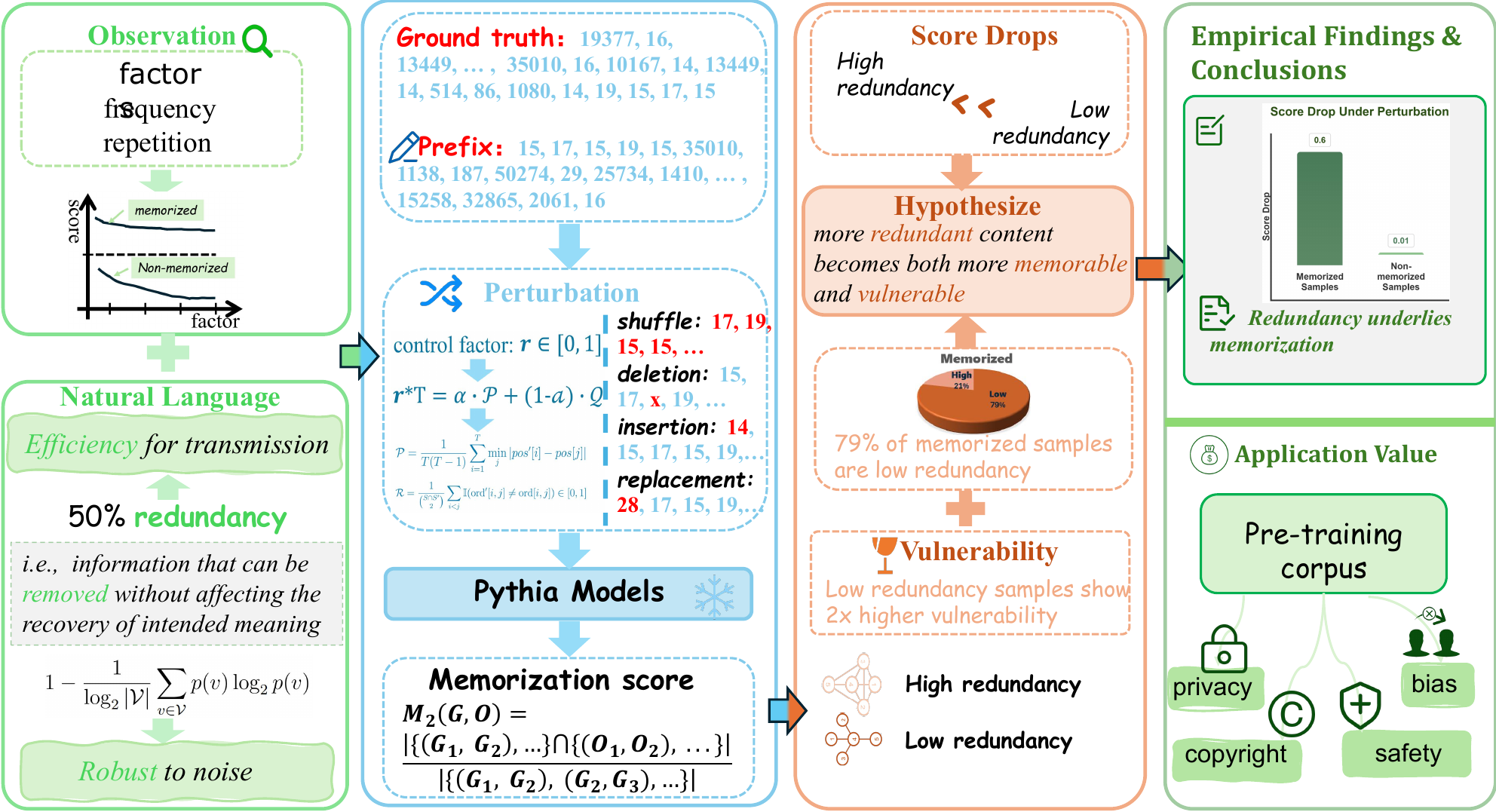}
  \caption{The framework of our method.}
  \label{fig:framework}
\end{figure}

To explore this disparity, we move beyond surface-level characteristics to investigate information redundancy as a more implicit factor influencing memorization. Our framework is shown in Figure~\ref{fig:framework}. 
Motivated by \citet{fedorenko2024language}'s insight that natural language maintains $~$50\% redundancy to balance information transmission efficiency and robustness to noise, we hypothesize that  redundancy underlies these differences. We developed a novel perturbation method that formalizes common text errors (insertion, deletion, replacement, shuffle), then unifies and  quantifies perturbation strength through relative and absolute positional changes of identical tokens.  We measured redundancy using entropy-based methods, where lower values indicate higher noise sensitivity.
Through comparative analysis across different categories (memorized vs non-memorized, high vs low redundancy), our results demonstrate: $\sim 79\%$ of memorized samples are low-redundancy, these low-redundancy samples exhibit 2-fold higher vulnerability than high-redundancy ones, and memorized samples lose 0.6 of their scores under perturbation while non-memorized samples drop only 0.01. 

Through  perturbation analysis across model scales, our contributions are:
\begin{itemize}
    \item Discovery that traditional memorization factors impact memorized and non-memorized samples differently, revealing that non-memorized samples are more sensitive to these factors' variation.
    \item Novel perturbation quantification method that enables unified quantified analysis across different perturbation types.
    \item Demonstration that   low-redundancy samples showing 2-fold higher fragility, and $\sim$79\% of memorized samples are low-redundancy, explaining inherent memorization vulnerability.
\end{itemize}

\section{Related Work}
Memorization research  has evolved along two complementary dimensions: understanding the underlying neural mechanisms (where memorization occurs) and identifying the factors that influence memorization behavior (what gets memorized and why).
The first line investigates the underlying neural mechanisms: early studies localized memorization within MLP neurons \citep{geva-etal-2023-dissecting, bayazit-etal-2024-discovering, zhu-etal-2024-fastmem} and specific layers \citep{haviv-etal-2023-understanding}, \citet{chang-etal-2024-localization} revealed distributed processing across layers, and recent work identified attention modules in deeper blocks as primary drivers \citep{menta-etal-2025-analyzing}.  \cite{huang-etal-2024-demystifying} challenged localized views, finding that memorization leverages distributed states and general language capabilities. 
{However, these mechanistic studies primarily focus on where memorization occurs rather than why certain content is preferentially memorized, leaving the underlying selection principles unexplored.}

In parallel, another research direction focuses on identifying the factors impact memorization.  Initial observations of elevated memorization in atypical samples \citep{NEURIPS2022_97011c64} evolved into  identification of key factors, including repetition, prefix length \citep{NEURIPS2023_7bc4f74e,274574,carlini2023quantifying,prashanth2025recite}. Deeper analysis revealed content hierarchies: models prioritize nouns and numbers as sample identifiers \citep{10.5555/3600270.3603043}, with special characters outperforming repeated tokens as memorization triggers \citep{bai2024specialcharactersattackscalable}. Scaling studies established predictable relationships, including linear scaling with model size, exponential with training epochs \citep{tirumala2022memorization,lu-etal-2024-scaling}, and log-linear patterns across capacity, repetition, and frequency \citep{carlini2023quantifying,NEURIPS2023_59404fb8}. 
{Yet these surface-level characteristics show inconsistent effects across different sample types, suggesting that deeper underlying factors may also impact memorization behavior.}

Building on these factor-based analyses,  recent research examining memorization under perturbations reveals complex sensitivity patterns that challenge intuitive expectations. While \citet{stoehr2024localizing} demonstrated robust memorization persistence under significant text modifications, \cite{xie2025memorizationlargelanguagemodels} showed fragility under slight perturbations in task-specific contexts, highlighting the nuanced nature of memorization robustness. 
Our work advanced this direction by introducing, to our knowledge, the first comparative analysis between memorized and non-memorized samples using unified perturbation quantification methods. Unlike prior work that treats all samples uniformly, our  differential analysis reveals that information redundancy influences memorization patterns, addressing the gap between surface-level observations and underlying mechanisms while  providing actionable guidance for improving training data quality, mitigating privacy risks and bias.

% % 这俩实验结果是相反的
% 【体现据我们所知，我们是第一个将memorized的样本和non-memorized样本分开研究的工作，并且发现了信息密度影响是memorization的因素，要体现发现信息密度影响memorization这一发现的应用价值。最后还要补充一个现在的研究memorization的大多数相关工作研究的是pythia】
% 介绍pythia，介绍与我们工作的区别

\section{Methodology}

\subsection{Unified Perturbation Quantification Strategy}\label{sec:perturbation_strategy}
Our perturbation strategy was motivated by the need to  quantify how information redundancy affects memorization  across diverse error types commonly encountered in real-world text processing.  Our perturbation strategy consisted of four steps: (1) cataloging common typos in human-generated text, (2) formalizing these errors mathematically, (3) developing unified quantification strategy, and (4) generating specific perturbation instances given perturbation type and intensity.

Firstly, we focused on four prevalent perturbation types that capture the full spectrum of structural and lexical modifications possible in text. Human-generated text frequently contains errors, which we categorize into four distinct types: token insertion (adding information), token deletion (removing information), token position shuffling (reorganizing information), and token substitution (altering information). These categories capture major types of text modifications, ensuring our analysis captures all possible ways that information structure can be disrupted.

Secondly, we formalized these perturbations to enable rigorous quantitative analysis across different modification types.  Specifically, we formalize perturbations as transformation operators $\Phi$ that map an original sequence $S = [s_1, s_2, \ldots, s_T]$ to a modified sequence $S'$. 
The \textbf{shuffle} operator $\Phi_{shuf}$ models role of sequential organization in memorization through $k$ sequential pairwise swaps, where each swap randomly selects two distinct positions $i$ and $j$ and exchanges their elements. This operation preserves lexical content while disrupting sequential organization, allowing us to isolate the contribution of positional relationships to memorization.  The \textbf{deletion} operator $\Phi_{del}$ simulates information loss by removing elements at randomly chosen positions $I_{del} \subset \{1, \ldots, T\}$ of cardinality $k$, yielding $S' = [s_i]_{i \in \{1, \dots, T\} \setminus I_{del}}$ with reduced length $T' = T - k$.
Conversely, the \textbf{insertion} operator $\Phi_{ins}$ introduces $k$ vocabulary-sampled tokens $T_{ins} = [t_1, \ldots, t_k]$ at random positions, expanding sequence length to $T' = T+k$. Finally, the \textbf{replacement} operator $\Phi_{rep}$ substitutes tokens at indices $I_{rep} \subset \{1, \ldots, T\}$ with vocabulary ($\mathcal{V}$) drawn alternatives, such that:
$$s'_i = \begin{cases} t_i \sim p(t|\mathcal{V}) & \text{if } i \in I_{rep} \\ s_i & \text{if } i \notin I_{rep} \end{cases}$$

Thirdly, we performed unified quantification of the aforementioned formalized perturbations. Specifically, our framework decomposes perturbation impact into two complementary dimensions: the absolution position change and the relative position change. The absolute position $\mathcal{P}$ quantifies the average normalized displacement of tokens from their original positions:
$$ \mathcal{P} = \frac{1}{T(T-1)} \sum_{i=1}^{T} \min_{j} |pos'[i] - pos[j]|$$
where $pos'[i]$ represents the position of token $i$ in the perturbed sequence, $pos[j]$ denotes the position of the corresponding token in the original sequence, and the minimum operation handles cases where tokens appear multiple times. The normalization by $T \cdot (T-1)$ ensures $\mathcal{P} \in [0,1]$.
The relative position $\mathcal{R}$ measures the preservation of pairwise ordering relationships between tokens:
$$\mathcal{R} = \frac{1}{\binom{S \cap S'}{2}} \sum_{i<j} \mathbb{I}(\text{ord}'[i, j] \neq \text{ord}[i, j]) \in [0, 1] $$
where $|S \cap S'|$ represents the cardinality of tokens present in both original and perturbed sequences, and the indicator function $\mathbb{I}(x)$ returns 1 when condition $x$ is satisfied (i.e., when the relative order of token pair $(i,j)$ differs between sequences) and 0 otherwise. The terms $\text{ord}'[i, j]$ and $\text{ord}[i, j]$ denote the relative ordering of tokens $i$ and $j$ in the perturbed and original sequences, respectively.
The unified perturbation magnitude $P$ combines these $\mathcal{P}, \mathcal{R}$ through weighted summation: 
$P = \alpha \cdot \mathcal{P} + (1-\alpha) \cdot \mathcal{R}$, 
where hyperparameter $\alpha \in [0,1]$ balances the relative importance of $\mathcal{P}$ and $\mathcal{R}$.

Finally, to facilitate the study of how different intensities of perturbation affect memorization, we employ an control factor $r\in [0, 1]$
to control the perturbation intensity of different types. Specifically, given a perturbation type and $r$, we first calculate the number of tokens to be modified as $r*T$, and randomly select the corresponding positions to be adjusted. Once the positions are determined, we perform insertion, deletion, substitution, and swapping operations according to the perturbation type to achieve the desired perturbation intensity.

% \vspace{4pt}
% \textbf{Redundancy.} Building on \citep{ZHAO2024122700}, which establishes varying word order dependence across tasks, we measure prefix redundancy as the entropy of a discrete sequence: \( \mathcal{A} = -\sum_{i} p_i \log_2 p_i \), where \( p_i \) represents the probability of the \( i \)-th unique element. Higher entropy indicates greater information content and lower redundancy \citep{fedorenko2024language}.

% \vspace{4pt}
% \textbf{Uncertainty.} To assess how prefix perturbations affect generated text, we measure model uncertainty as the entropy of the predicted probability distribution: \( \mathcal{U} = -\sum_{j} q_i \log_2 q_j \), where \( q_j \) denotes probabilities derived from the model's logits. This entropy reflects the dispersion of predicted probabilities, indicating the model's generation confidence.

% 缺一些逻辑连贯的句子
\subsection{Prefix Information Redundancy}

Redundancy refers to information that can be removed without affecting the recovery of intended meaning. Conceptually, redundancy measures the predictability of text elements: high redundancy indicates that many tokens can be predicted from context, while low redundancy indicates that most tokens carry unique, unpredictable information.   Intuitively, natural language typically exhibits higher redundancy, representing a trade-off between communication efficiency and noise sensitivity, while academic papers employ more concise language with stronger logical structure, resulting in lower redundancy.  This distinction is crucial because memorization mechanisms may preferentially target content with specific redundancy profiles. We employed the method introduced in \cite{fedorenko2024language} for measurement. Given a sequence $S$, the calculation formula is:
\begin{equation}
Re(s) = 1 - \frac{1}{\log_2 |\mathcal{V}|} \sum_{v \in \mathcal{V}} p(v) \log_2 p(v)
\end{equation}

\subsection{Memorization Quantification Metric}

Commonly used memorization evaluation relies on exact unigram matching \citep{carlini2021extracting, carlini2023quantifying, NEURIPS2023_59404fb8, prashanth2025recite},  recent work has adopted ROUGE-based frameworks  to assess overlapping n-gram subsequences \citep{luo2025sharedpathunravelingmemorization}. We also employ this latter metric.  Specifically, given a generated output $\text{O} = [o_1, o_2, ..., o_l]$ and its corresponding ground truth instance $\text{G} = [g_1, g_2, ..., g_m]$, the memorization score is computed as:
\begin{equation}
\mathcal{M}_n(\text{O},\text{G}) = \frac{|\mathcal{N}_n(\text{O}) \cap \mathcal{N}_n(\text{G})|}{|\mathcal{N}_n(\text{G})|} \in [0, 1]
\end{equation}

Here, $\mathcal{N}_n(\text{G}) = \{(G_i, G_{i+1}, ..., G_{i+n-1}) : 1 \leq i \leq |G| - n + 1\}$ denotes the set of all consecutive $n$-gram tokens within sequence $\text{S}$. This metric captures memorization ranging from individual token recall to substantial sequence reproduction. We establish memorization through the criterion $\mathcal{M}_n > \theta$, where $\theta$ serves as the classification threshold.

% \paragraph{\textcolor{lise}{Memorization Efficiency}} To further explore the correlation between model dynamics (scale and trained steps) and memorization, we propose  \textit{memorization efficiency} to measure the per-parameter memorization. Given a test set $\text{D} = \{x_1, x_2, \dots, x_N\}$, we define memorization efficiency as the ratio between memorized samples and total parameter count:
% \[
% e = \frac{\sum_{x_i \in \text{D}} \mathbb{I} \big( \mathcal{M}_{n}(x_i, x_{i}{'}) > 0.5 \big)}{\sum_{l=1}^{\text{L}-1} \sum_{i=1}^{l} \sum_{j=1}^{l+1} |w^{l}_{ij}| + \sum_{l=1}^{\text{L}} \sum_{i=1}^{n^{l}} |b^{l}_i|},
% \]
% where $\mathbb{I}(\cdot)$ represents the indicator function, $\text{L}$ denotes the number of model layers, $n^{l}$ signifies neurons in layer $l$, and $w^{l}_{ij}$ and $b^{l}_i$ correspond to weights and biases, respectively. The result is shown in Figure \ref{subfig:memory_efficiency}.

\section{Experimental Setup}

\paragraph{\textcolor{lise}{Model Selection and Rationale}}
We selected the Pythia model family, which is ideal for memorization  due to its open access to model weights, pretraining data, and training checkpoints. This family includes eight models of varying scales (70M to 12B parameters), all trained on The Pile corpus \citep{gao2020pile} with identical configurations, and the availability of checkpoints spanning the entire pre-training process. This unique feature enables a fine-grained analysis of memorization which has been used by \citet{10.5555/3618408.3618510,  gurnee2023finding, huang-etal-2024-demystifying,lesci-etal-2024-causal, prashanth2025recite}.
We evaluated all eight scales of Pythia models across different training checkpoints, including 20K, 40K, 60K, 80K, 100K, and 120K steps, allowing us to observe how memorization behavior develops as training progresses. This level of transparency is unavailable for many other popular LLMs like LLaMA, Mistral, and Deepseek, making them unsuitable for this study. While OLMo models also provide open-source pretraining corpora, they do not offer checkpoints at different training steps \citep{srivastava2025owlprobingcrosslingualrecall}.

\paragraph{\textcolor{lise}{Test Dataset Construction and Validation}}
Testing memorization requires using pretraining corpora as test datasets, but their massive size makes full testing impractical. We can only test subsets, yet even preprocessing these subsets is computationally expensive. To enable more efficient memorization testing, we utilized the test dataset released by \cite{NEURIPS2023_59404fb8}. This dataset comprises 32-token extractable sequences sourced from The Pile corpus. Each sample contains a 64-token sequence partitioned into a 32-token prefix  and a 32-token continuation. To balance computational feasibility with statistical rigor, we selected approximately 10,000 sequences uniformly from the available test set, providing sufficient power for our differential analysis while enabling  evaluation across multiple model scales and perturbation types.

\begin{figure}[ht]
    \centering
    % \small
    \begin{subfigure}[t]{0.32\linewidth}
        \centering
        \includegraphics[width=\linewidth]{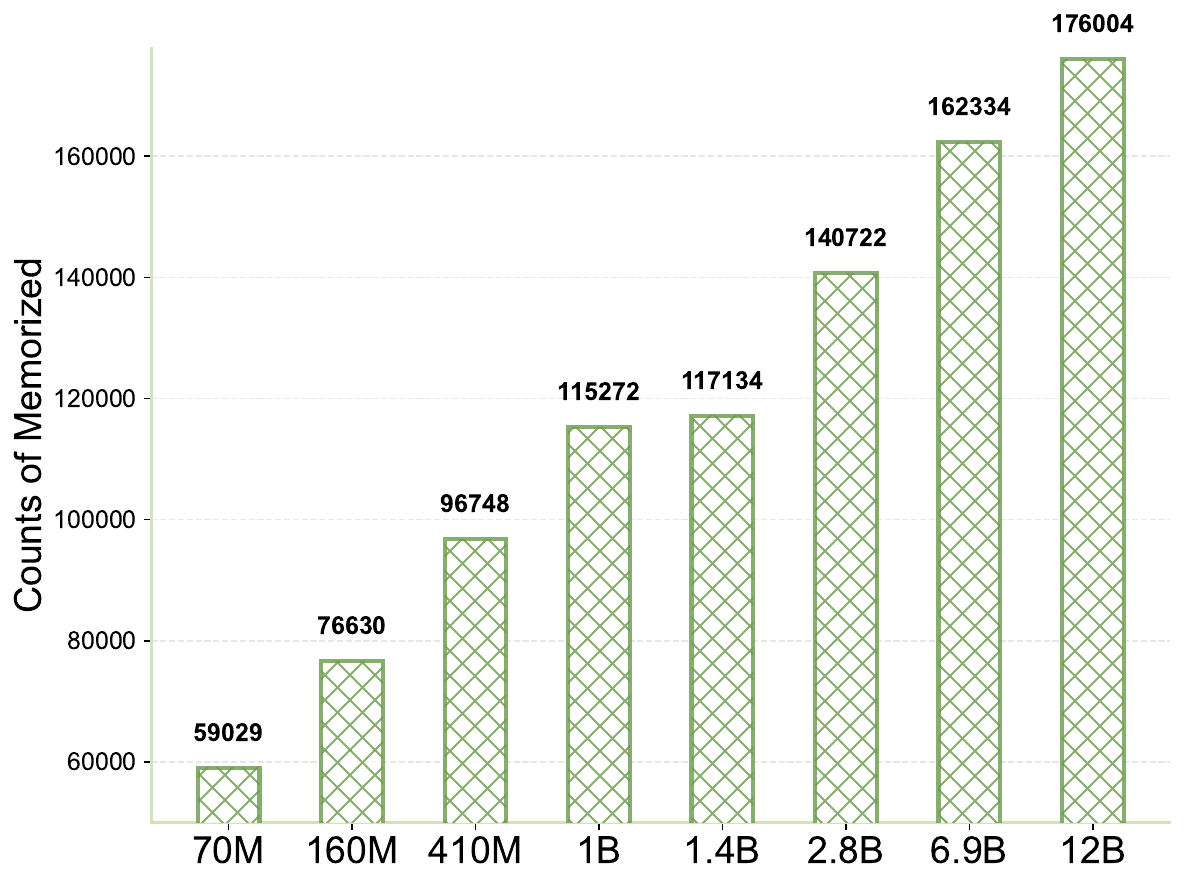}
        \caption{Memorized counts of dataset in \citep{NEURIPS2023_59404fb8}}
        \label{subfig:memorized_counts}
    \end{subfigure}
    % \hfill
    % \vspace{8pt}
    \begin{subfigure}[t]{0.32\linewidth}
        \centering
        \includegraphics[width=\linewidth]{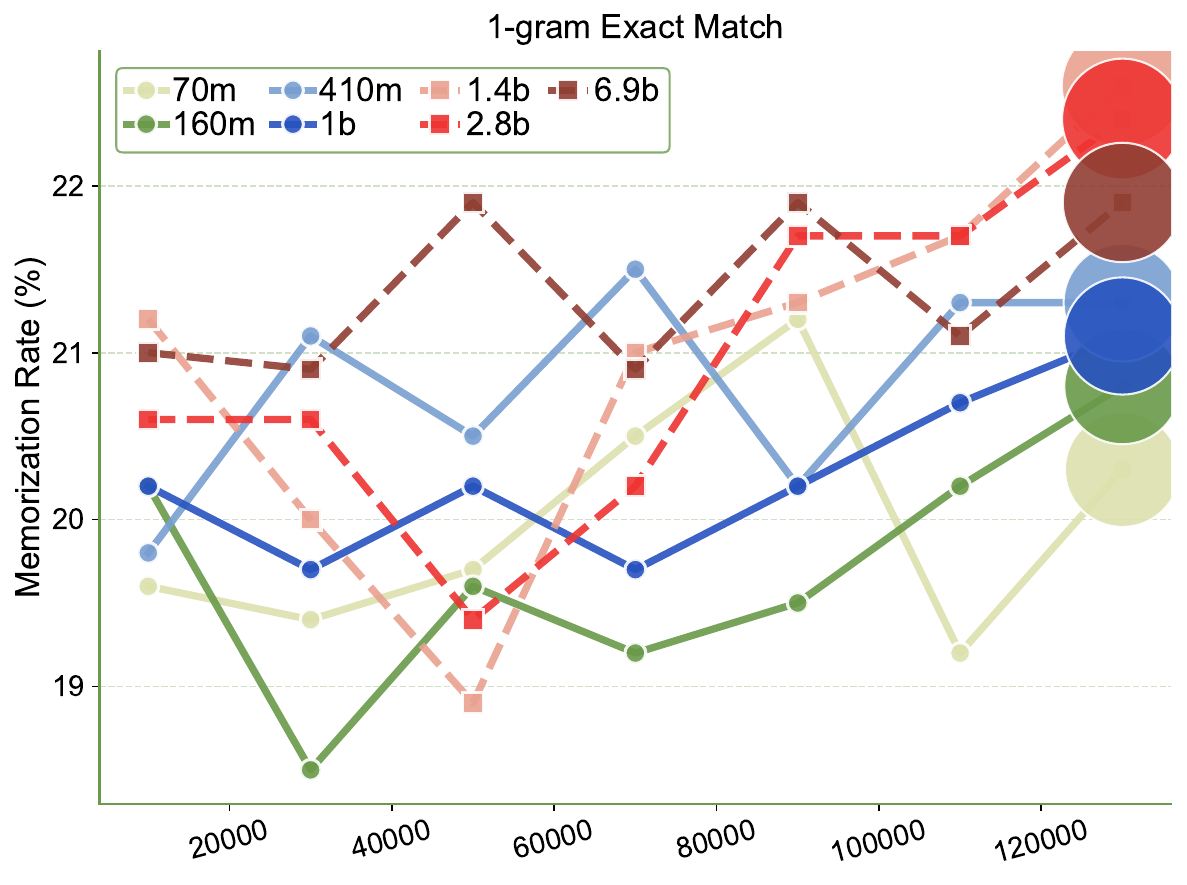}
        \caption{Memorized rates in selected data (1-gram)}
        \label{subfig:mem_checkpoint_1gram}
    \end{subfigure}
    % \hfill
    % \vspace{8pt}
    \begin{subfigure}[t]{0.32\linewidth}
        \centering
        \includegraphics[width=\linewidth]{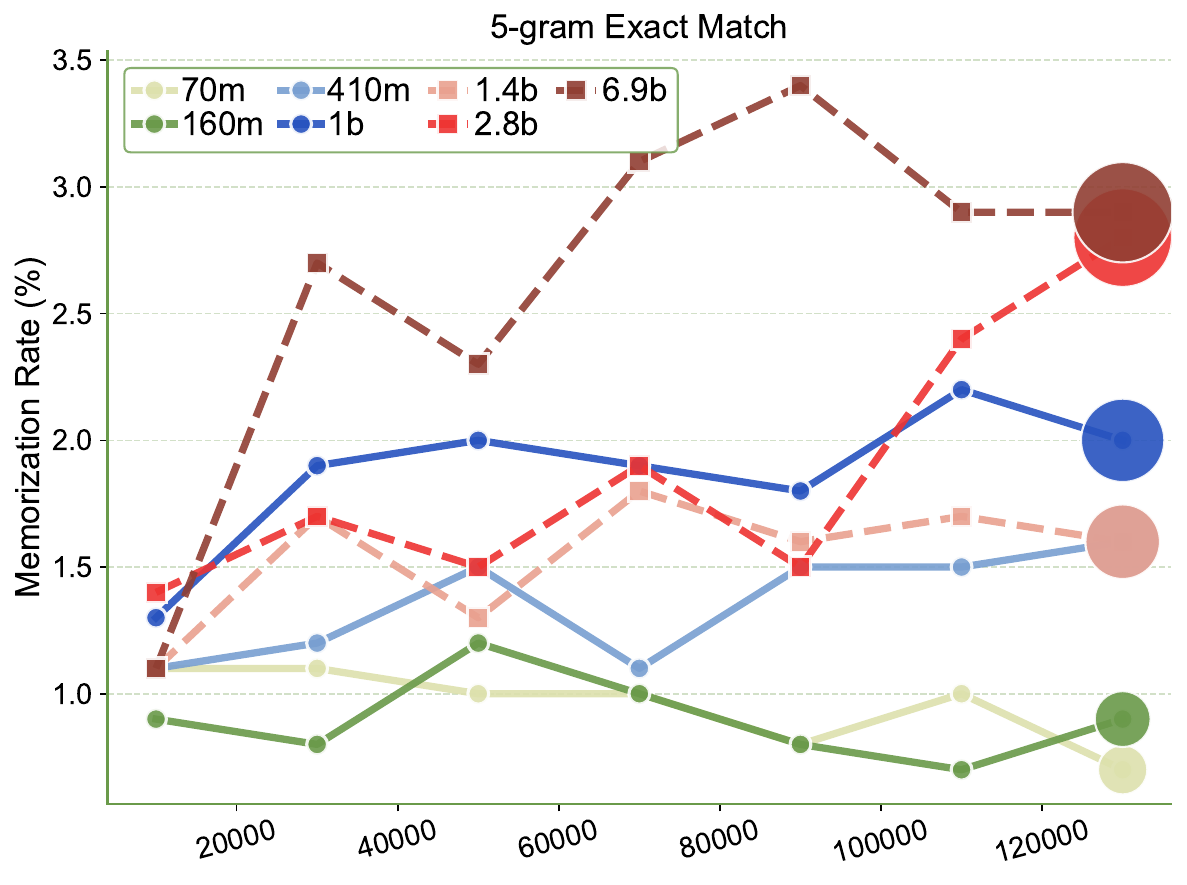}
        \caption{Memorized rates in selected data (5-gram)}
        \label{subfig:mem_checkpoint_5gram}
    \end{subfigure}
    % \hfill
    % \vspace{15pt}
    % \begin{subfigure}[t]{0.4\linewidth}
    %     \centering
    %     \includegraphics[width=\linewidth]{memory_efficiency_multi_ngram.pdf}
    %     \caption{Efficiency}
    %     \label{subfig:memory_efficiency}
    % \end{subfigure}
    \caption{Memorization scaling validation. Pilot study (N=1,000) demonstrates consistent scaling patterns with larger datasets, validating our sampling approach and confirming adequate statistical power for main analyses.}
    \label{fig:pilot_study}
\end{figure}

To validate the representativeness of our sampling approach, we conducted a pilot study using 1,000 sequences across different model scales and training checkpoints. As shown in Figure \ref{fig:pilot_study}, the pilot study with 1,000 sequences successfully captured the same scaling relationships (e.g., 3$\times$ increase from 70M to 12B models) and convergence patterns (19-22\% memorization rates) observed in larger datasets. This consistency validates our choice of 10,000 sequences as providing adequate statistical power for detecting differential memorization effects. 

\paragraph{\textcolor{lise}{Experimental Configuration}}
All experiments were conducted using greedy decoding to ensure deterministic and reproducible results, eliminating the need for multiple runs with different random seeds. We deployed our experiments on 1-2 NVIDIA A100 GPUs with 80GB memory each, providing sufficient computational resources for large-scale model evaluation. Following established practices in \citep{carlini2023quantifying}, we set the  $\theta$ to the default value of 0.5. For $n$-gram analysis, we used $n=2$ as the default setting unless explicitly specified otherwise. We  tested different values of the control factor $r \in [0,1]$, specifically evaluating $r\in \{1/16,1/8,1/2, 15/16\}$ to capture varying degrees of perturbation. The generation length was fixed at 32 tokens to match the continuation length in our test dataset, ensuring consistent evaluation across all experimental conditions.

\section{\textbf{Empirical Observation}}

\subsection{\textbf{Contrasting Effects on Memorized and Non-memorized Ones}}
\begin{figure}[!t]
    \centering

    \begin{subfigure}[b]{4.5cm}
        \centering
        \includegraphics[width=4.5cm,height=3.8cm]{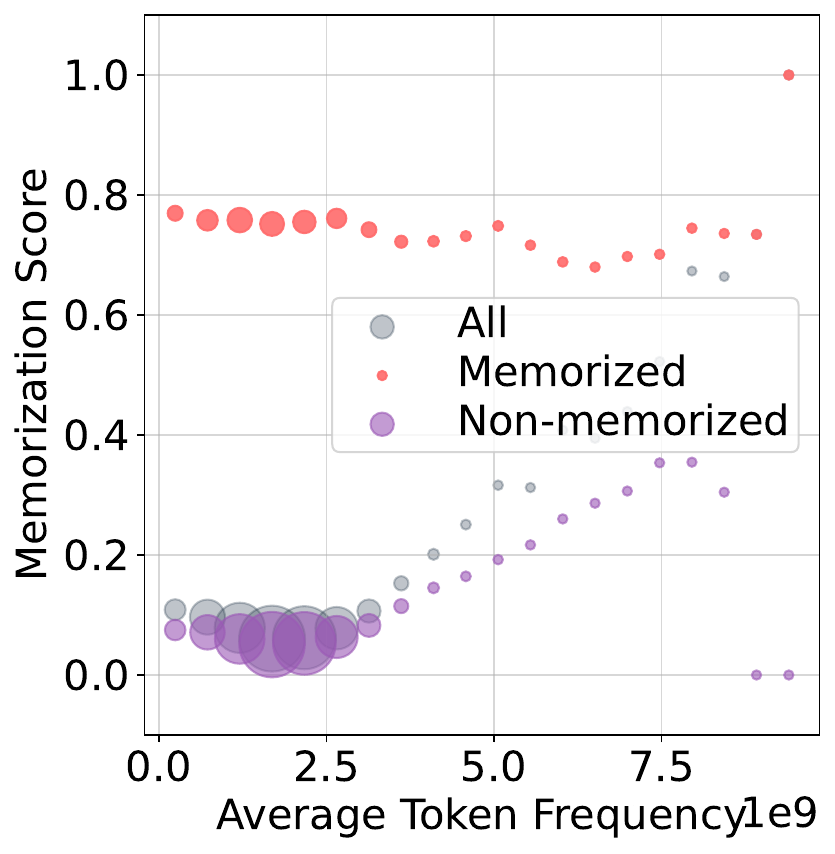}
    \end{subfigure}
    \begin{subfigure}[b]{4.0cm}
        \centering
        \includegraphics[width=4.0cm,height=3.8cm]{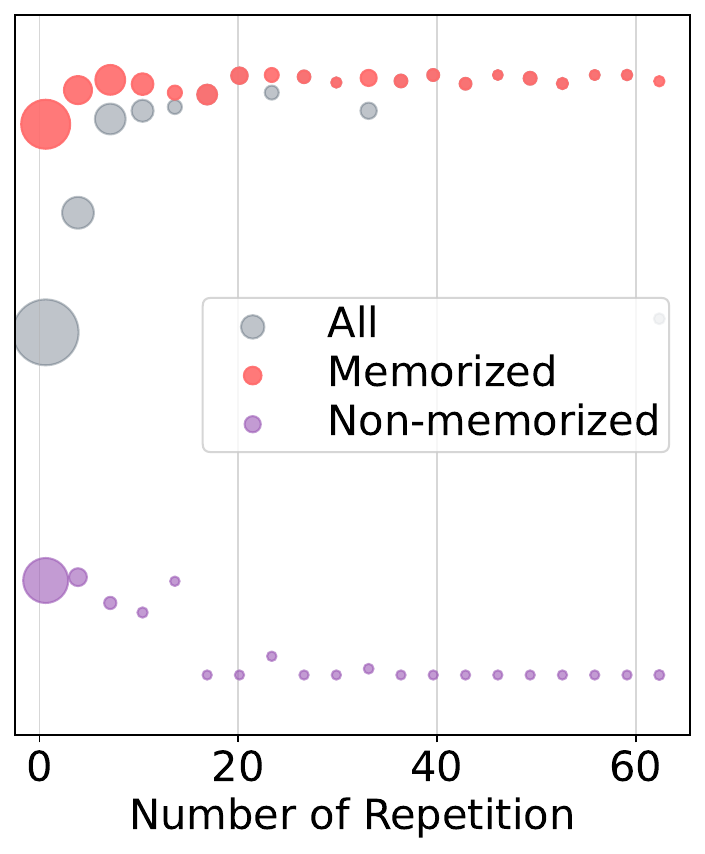}
    \end{subfigure}
    \begin{subfigure}[b]{4.0cm}
        \centering
        \includegraphics[width=4.0cm,height=3.8cm]{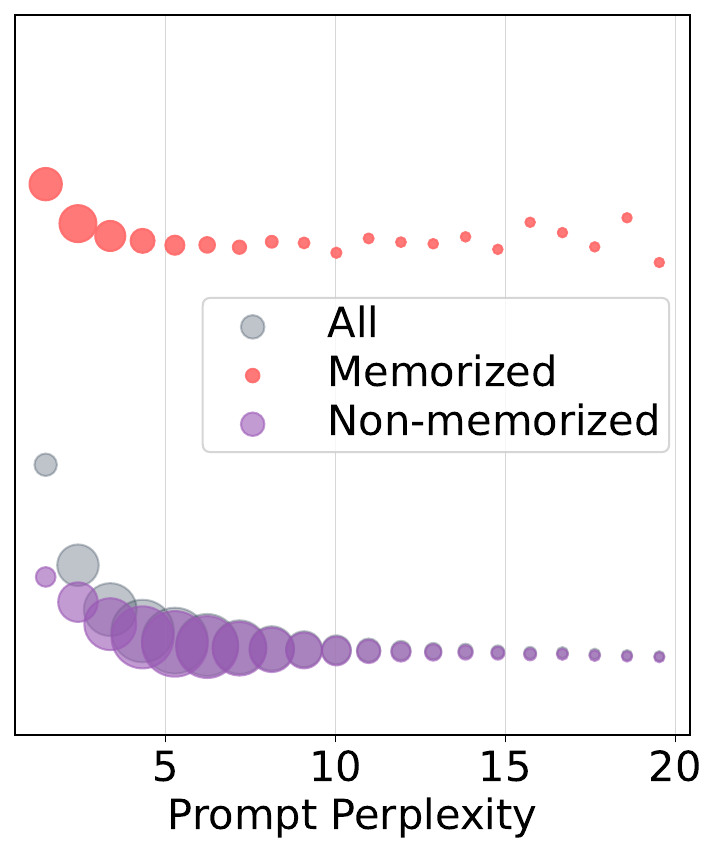}
    \end{subfigure}

    \begin{subfigure}[b]{4.5cm}
        \centering
        \includegraphics[width=4.5cm, height=3.8cm]{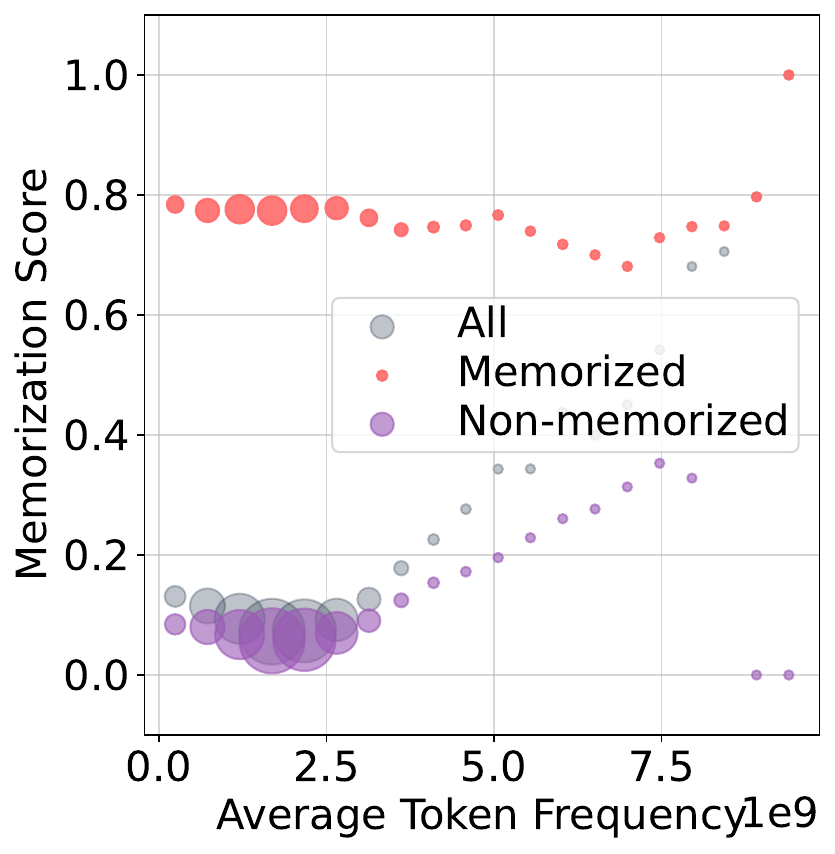}
    \end{subfigure}
    \begin{subfigure}[b]{4.0cm}
        \centering
        \includegraphics[width=4.0cm,height=3.8cm]{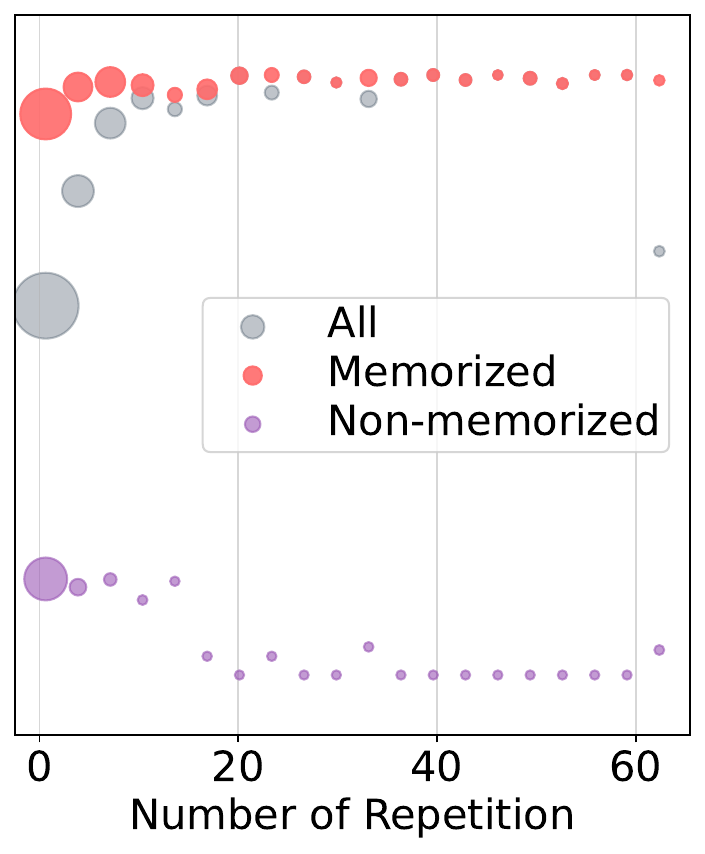}
    \end{subfigure}
    \begin{subfigure}[b]{4.0cm}
        \centering
        \includegraphics[width=4.0cm,height=3.8cm]{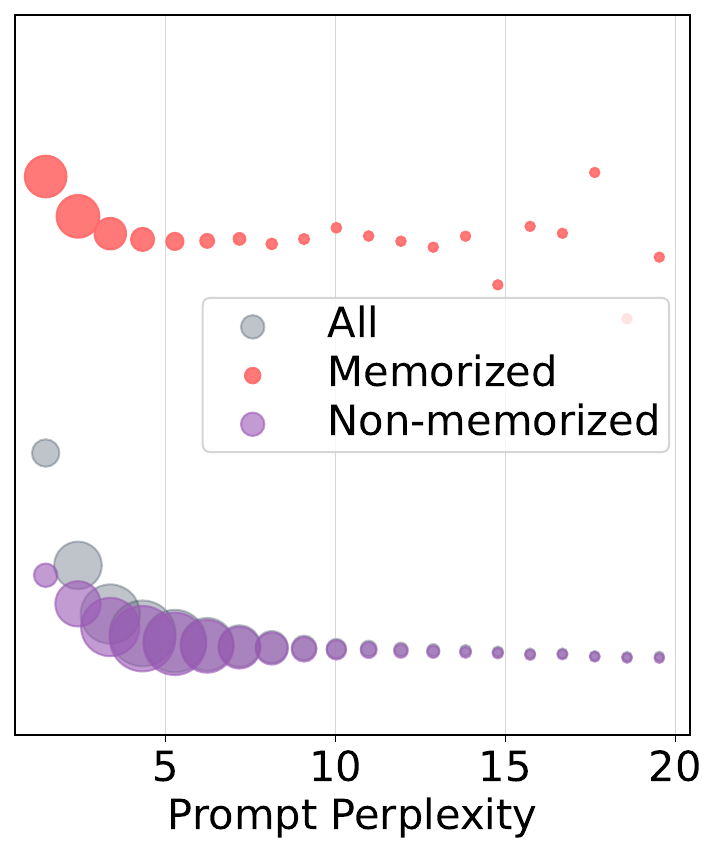}
    \end{subfigure}
    % \begin{subfigure}[b]{0.185\linewidth}
    %     \centering
    %     \includegraphics[width=\linewidth]{figures/2.8b/huffman_coding_length_analysis.pdf}
    % \end{subfigure}
    % \begin{subfigure}[b]{0.22\linewidth}
    %     \centering
    %     \includegraphics[width=\linewidth]{figures/2.8b/sequence_entropy_analysis.pdf}
    % \end{subfigure}

    % \begin{subfigure}[b]{0.22\linewidth}
    %     \centering
    %     \includegraphics[width=\linewidth]{figures/6.9b/avg_frequency_analysis.pdf}
    % \end{subfigure}
    % \begin{subfigure}[b]{0.185\linewidth}
    %     \centering
    %     \includegraphics[width=\linewidth]{figures/6.9b/num_repeating_analysis.pdf}
    % \end{subfigure}
    % \begin{subfigure}[b]{0.186\linewidth}
    %     \centering
    %     \includegraphics[width=\linewidth]{figures/6.9b/prompt_perplexity_analysis.pdf}
    % \end{subfigure}
    % \begin{subfigure}[b]{0.185\linewidth}
    %     \centering
    %     \includegraphics[width=\linewidth]{figures/6.9b/huffman_coding_length_analysis.pdf}
    % \end{subfigure}
    % \begin{subfigure}[b]{0.185\linewidth}
    %     \centering
    %     \includegraphics[width=\linewidth]{figures/6.9b/sequence_entropy_analysis.pdf}
    % \end{subfigure}

    \begin{subfigure}[b]{4.5cm}
        \centering
        \includegraphics[width=4.5cm, height=3.8cm]{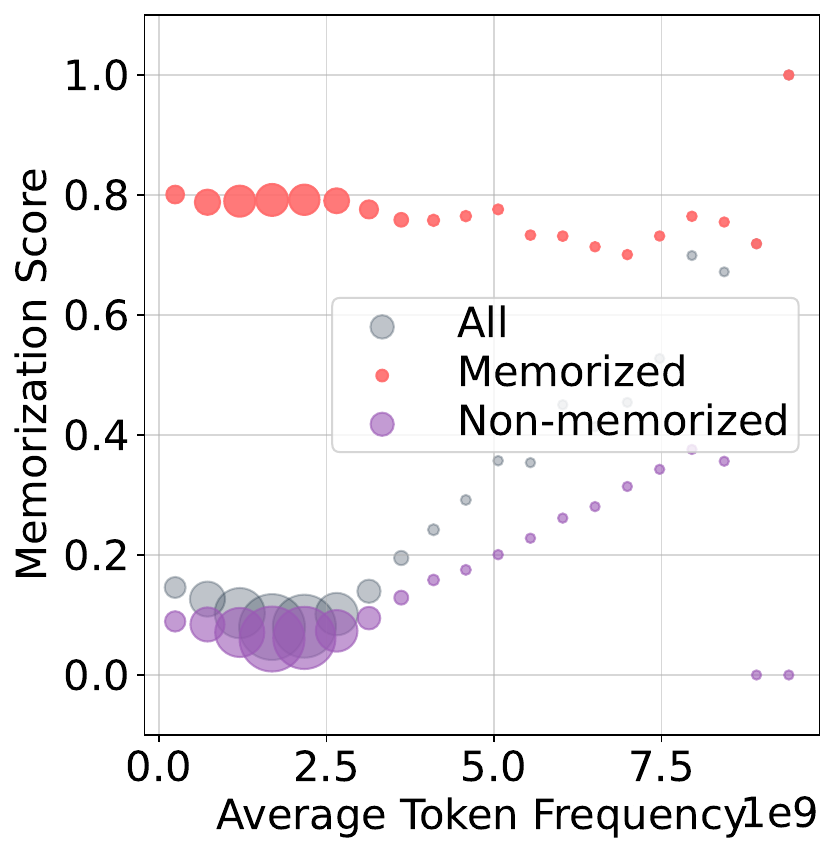}
        \caption{Token frequency}
        \label{fig:avg_frequency}
    \end{subfigure}
    \begin{subfigure}[b]{4.0cm}
        \centering
        \includegraphics[width=4.0cm, height=3.8cm]{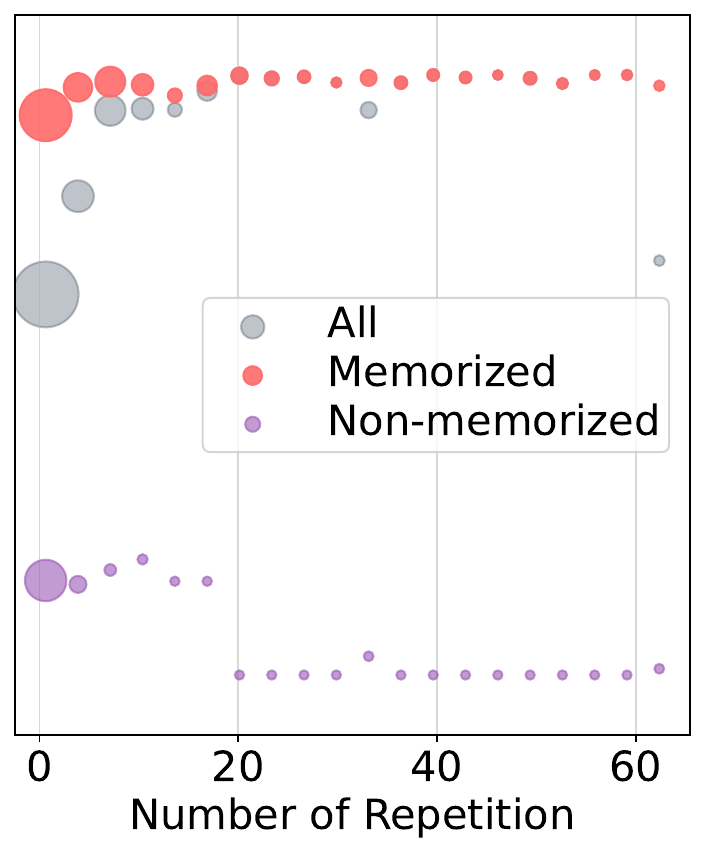}
        \caption{Repetitions}
        \label{fig:num_repeating}
    \end{subfigure}
    \begin{subfigure}[b]{4.0cm}
        \centering
        \includegraphics[width=4.0cm, height=3.8cm]{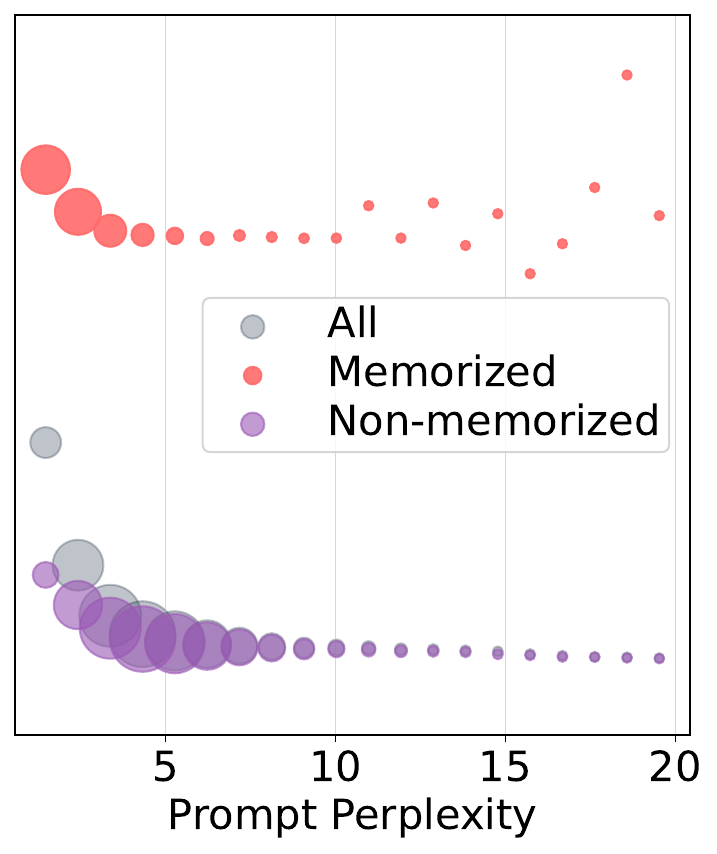}
        \caption{Perplexity}
        \label{fig:prompt_perplexity}
    \end{subfigure}
    % \begin{subfigure}[b]{0.185\linewidth}
    %     \centering
    %     \includegraphics[width=\linewidth]{figures/12b/huffman_coding_length_analysis.pdf}
    %     \caption{Compressibility}
    %     \label{fig:compressibility}
    % \end{subfigure}
    % \begin{subfigure}[b]{0.22\linewidth}
    %     \centering
    %     \includegraphics[width=\linewidth]{figures/12b/sequence_entropy_analysis.pdf}
    %     \caption{Redundancy}
    %     \label{fig:Redundancy}
    % \end{subfigure}
    \vspace{-3mm}
    \caption{Data characteristic effects on memorization scores across Pythia 410M, 6.9B, and 12B models (n=1).} % under 1-gram
    \label{fig:exp_2_1_RQ_3}
\end{figure}

Prior work shows that higher token frequency, greater repetition count, and lower prompt perplexity increase memorization probability \citep{carlini2021extracting, carlini2023quantifying, prashanth2025recite}. 
 However, these findings stem from aggregate correlation analyses across all samples. By separately examining memorized versus non-memorized groups, we uncovered distinct behavioral patterns that remain hidden in aggregate statistics, even under identical data characteristic values. Figures \ref{fig:avg_frequency}-\ref{fig:num_repeating} illustrate these patterns, with ascending data characteristic values (x-axis) versus 2-gram memorization scores (y-axis). 
A clear trend emerges: data characteristic values exert substantial effects on  the non-memorized group, whereas their impact on the memorized group remains relatively minimal. This pattern holds consistently across all model scales, but we present representative findings from Pythia 410M, 2.8B, and 12B models due to space constraints.

Specifically, examining Pythia 12B results (bottom row), when average token frequency increases from 0 to 8 (Figure \ref{fig:avg_frequency}), memorized samples demonstrate stability with memorization scores declining minimally from 0.80 to 0.71 (decrease of 0.09). Non-memorized samples, however, exhibit a two-tiered response: scores first decrease from 0.15 to 0.05 as frequency rises from 0 to 2.5, then increase from 0.05 to 0.4 as frequency continues from 2.5 to 8. Similarly, when repetition count increases from 0 to 60 (Figure \ref{fig:num_repeating}), memorized groups show minimal change with scores rising slightly from 0.95 to 0.99 (modest increase of 0.04). Non-memorized groups again display threshold behavior: scores remain stable from 0 to 18 repetitions, then drop sharply from 0.18 to 0 as repetitions increase from 18 to 60, remaining constant thereafter. 
To statistically validate these differential patterns, we divided memorized and non-memorized samples into 10 equal-sized groups, calculated  memorization score variance  for each group, and performed $t$-tests between the resulting distributions. Token frequency and repetition count showed significant differences with $p < 0.001$, confirming that \textit{memorized samples are significantly less responsive to data characteristic variations than non-memorized samples}. 
Although perplexity universally impacts both groups with increasing values causing sharp memorization declines, this finding underscores the importance of stratified analysis for understanding memorization mechanisms.

\subsection{\textbf{Towards Understanding Underlying Factors}}
The differential effects observed above suggest that surface-level characteristics alone may not fully explain memorization behavior. The distinct sensitivity patterns between memorized and non-memorized samples indicate that deeper factors might be at play.
Motivated by \citet{fedorenko2024language}'s finding that natural language maintains $~$50\% redundancy to balance communication efficiency and noise robustness, we conjecture that information redundancy may be related to the observed differential patterns. Specifically, we hypothesize that memorized and non-memorized samples might differ in their redundancy profiles. To verify this, we develop a perturbation-based analysis method.

% potentially contributing to different vulnerability patterns under perturbation.

% 这里redundancy的逻辑怎么加进来？也没做实验分析

\section{\textbf{Unified Perturbation Method Validation}} 
Our unified perturbation quantification method (Section \ref{sec:perturbation_strategy}) operates on input sequences by measuring  positional disruptions. To validate that this input-side quantification accurately reflects actual model disruption, we examine whether equivalent perturbation strengths produce comparable output-side effects across different modification types.  We measure the correlation between perturbation strength ($r \times T$) and output uncertainty to verify that different perturbations with identical strength $r$ produce equivalent effects on model outputs. Uncertainty is quantified as the Shannon entropy $H = -\sum_i p_i \log p_i$ of the next-token probability distribution, where $p_i$ represents the output probability of the $i$-th \textit{generated} token. 
We select uncertainty as our validation metric because effective perturbations should theoretically disrupt learned patterns and increase prediction uncertainty. This enables us to test whether equivalent perturbation strengths produce comparable model disruption across different modification types. If our unified framework is valid, then perturbations with identical strength should yield similar uncertainty increases regardless of whether they involve shuffling, deletion, insertion, or replacement, confirming that we capture true disruptive impact rather than superficial input changes.
\begin{figure}[!t]
    \centering
    \begin{subfigure}{\linewidth}
        \centering
        \includegraphics[width=0.24\linewidth]{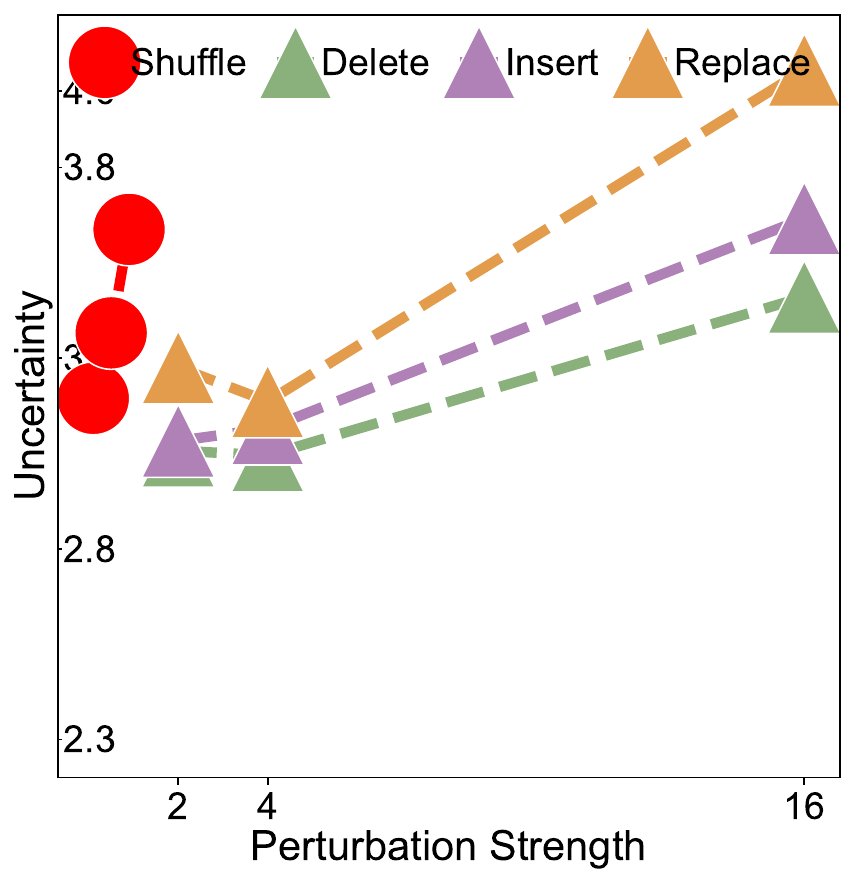}
        \includegraphics[width=0.24\linewidth]{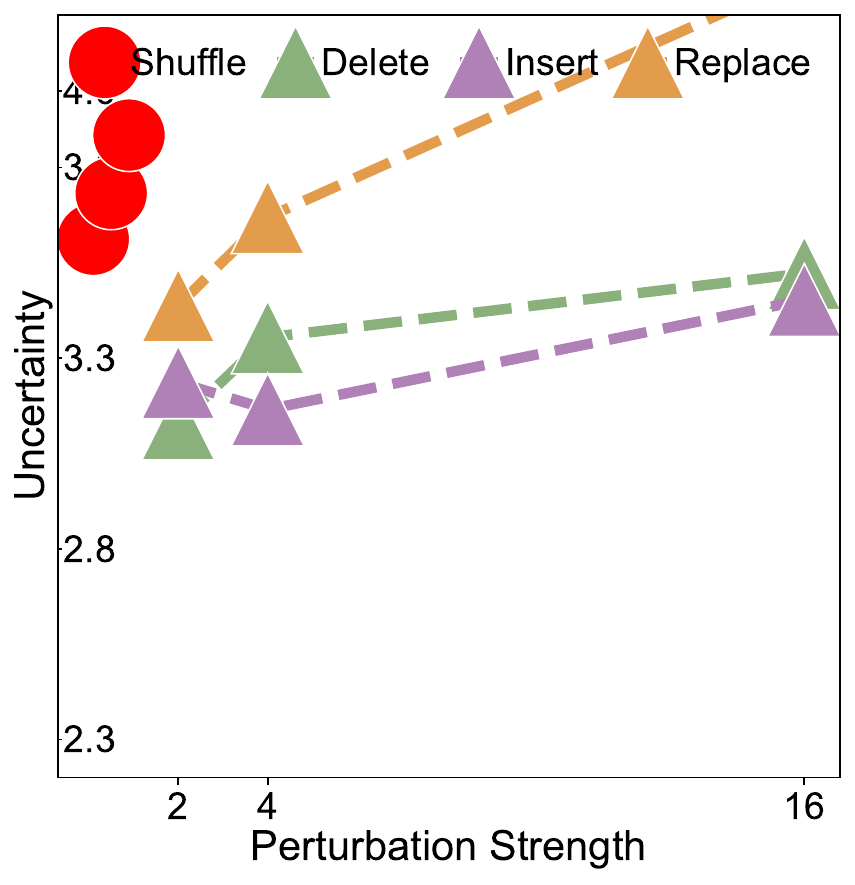}
        \includegraphics[width=0.24\linewidth]{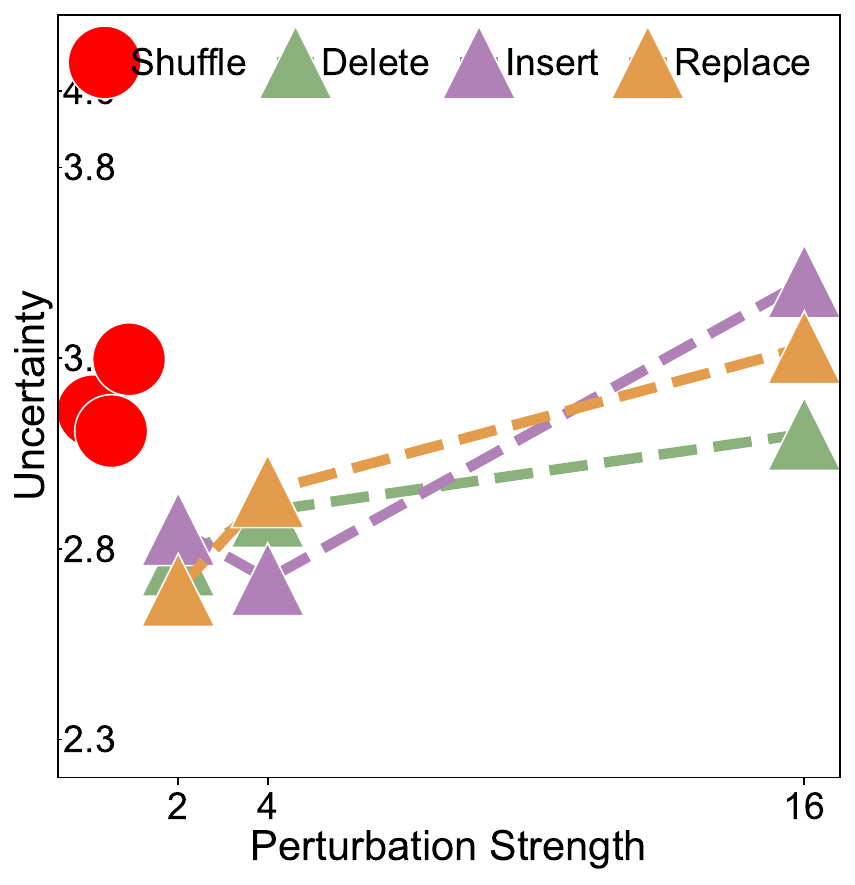}
        \includegraphics[width=0.24\linewidth]{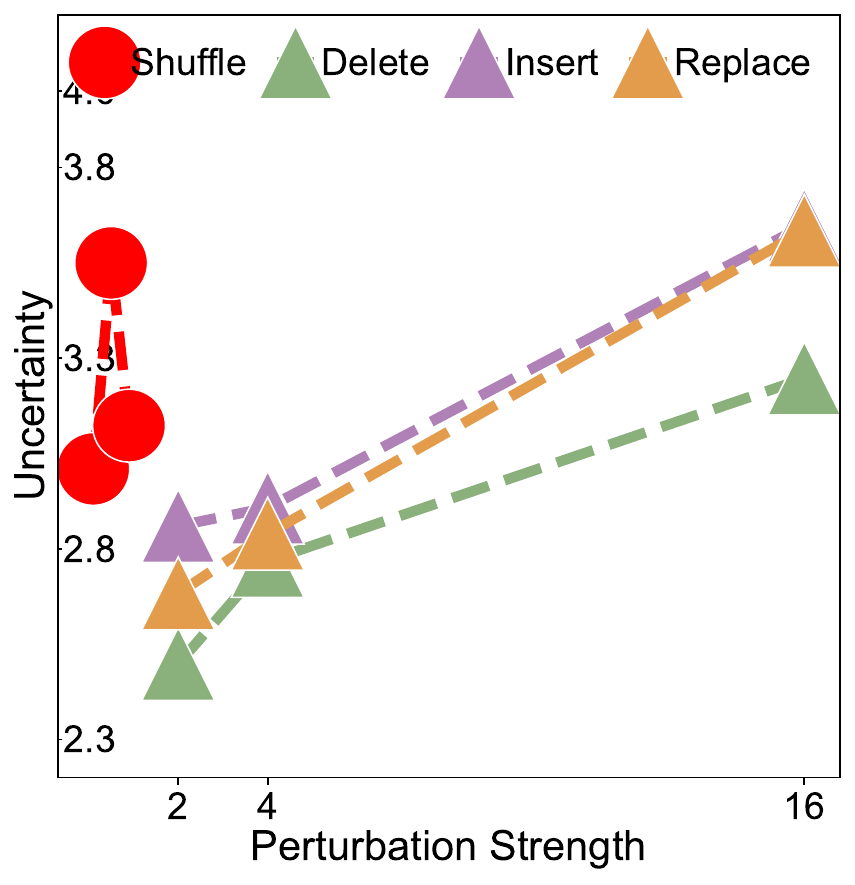}
        \includegraphics[width=0.24\linewidth]{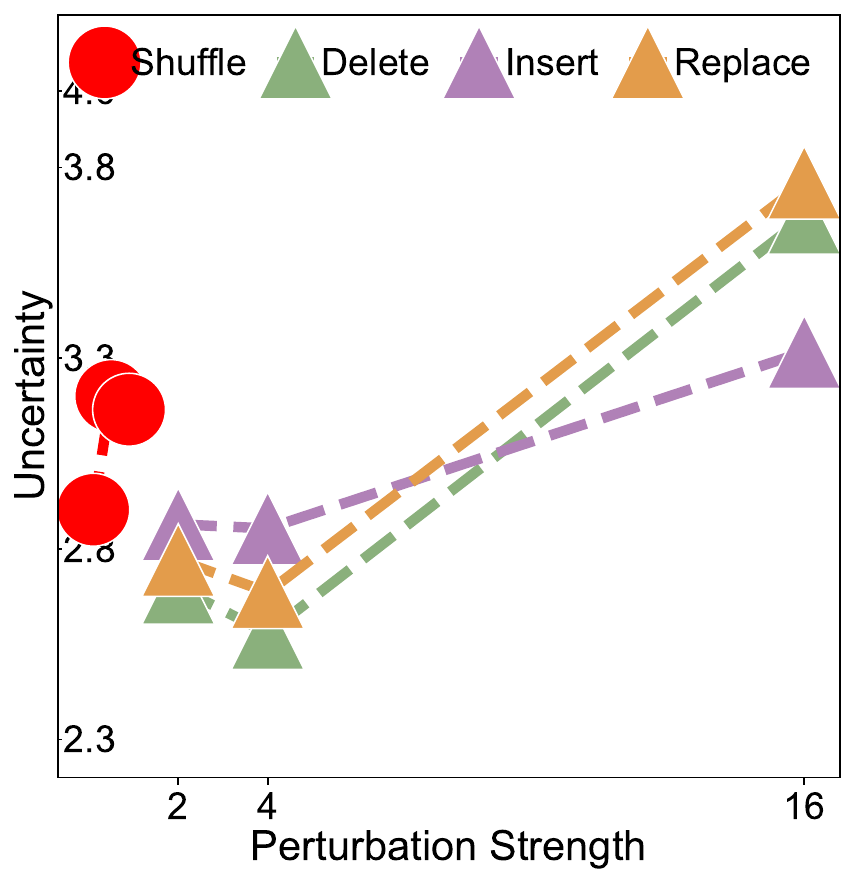}
        \includegraphics[width=0.24\linewidth]{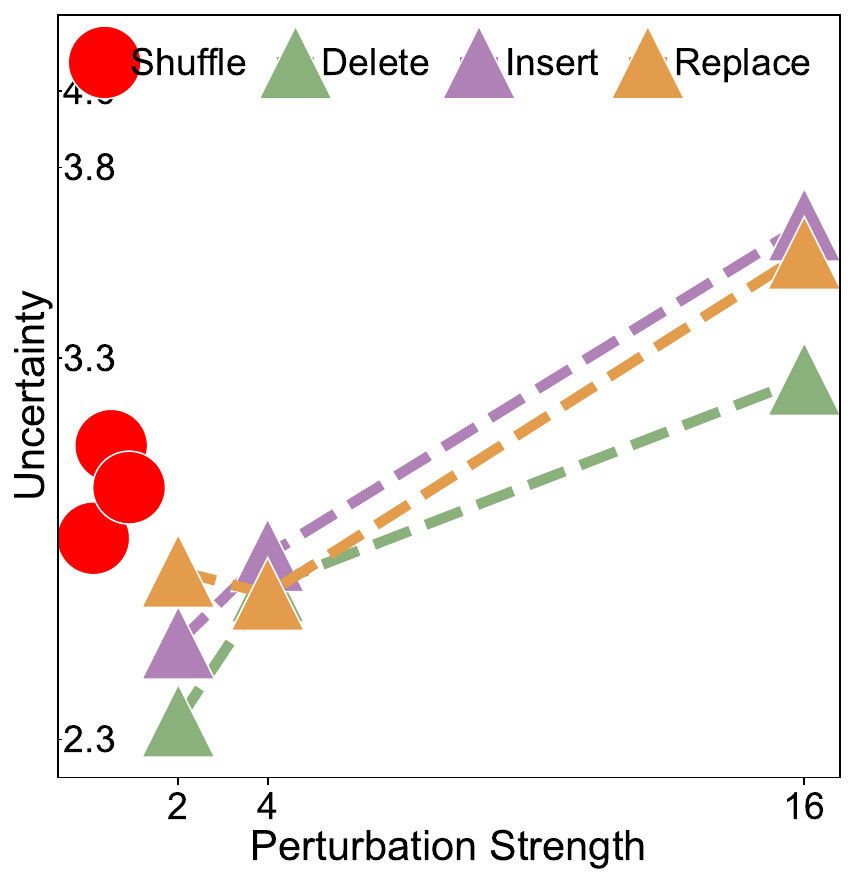}
        \includegraphics[width=0.24\linewidth]{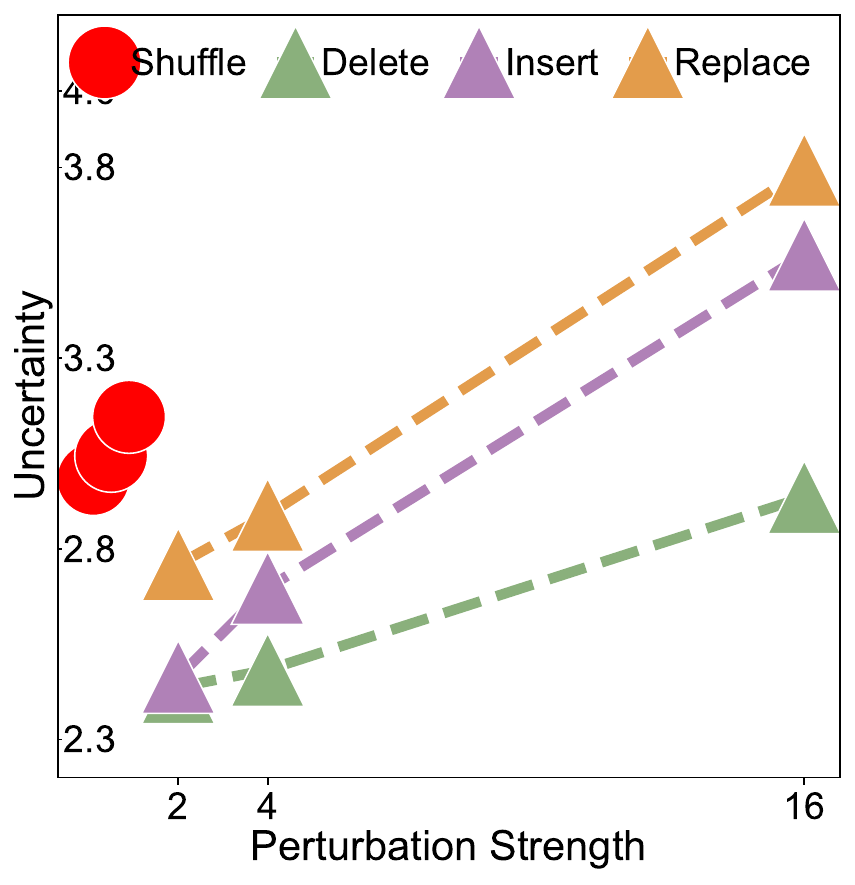}
        \includegraphics[width=0.24\linewidth]{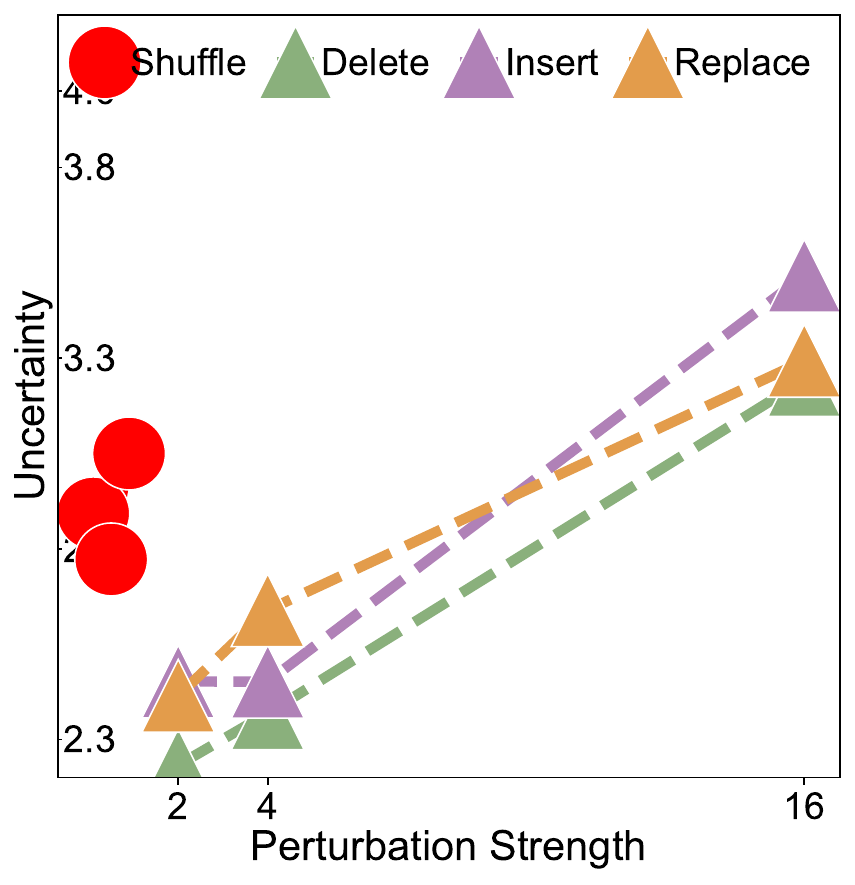}
    \end{subfigure}
    \caption{Impacts of perturbation strength on output uncertainty across Pythia models (70M, 160M, 410M, 1B, 1.4B, 2.8B, 6.9B, and 12B parameters).}
    \label{fig:perturbation_uncertainty}
\end{figure}

Figure \ref{fig:perturbation_uncertainty}  demonstrates strong positive linear correlations between perturbation strength and output uncertainty across all model scales, with uncertainty consistently increasing as perturbation intensity grows. 
Specifically, take Pythia 12B as an example (bottom right), as perturbation strength increases from 2 to 16 (i.e., $r$ increases from $1/16$ to $1/2$), uncertainty exhibits monotonic increase: shuffle method (2.3→3.8), deletion (2.4→3.7), insertion (2.5→3.6), and replacement (2.6→3.5), with coefficient of determination $R^2$ ranging from 0.82 to 0.91 (mean = 0.85) for all 32 model-perturbation combinations (4 types × 8 models), indicating that over 85\% of uncertainty variance is explained by the linear relationship with perturbation strength.  This confirms that equivalent strength values produce nearly identical model disruption across distinct perturbation types, thus validating our quantification framework for unified measurement of perturbation intensity to quantitatively analyze the relationship between perturbation strength and memorization.

\begin{figure}[!t]
    \centering

        \includegraphics[width=0.24\linewidth]{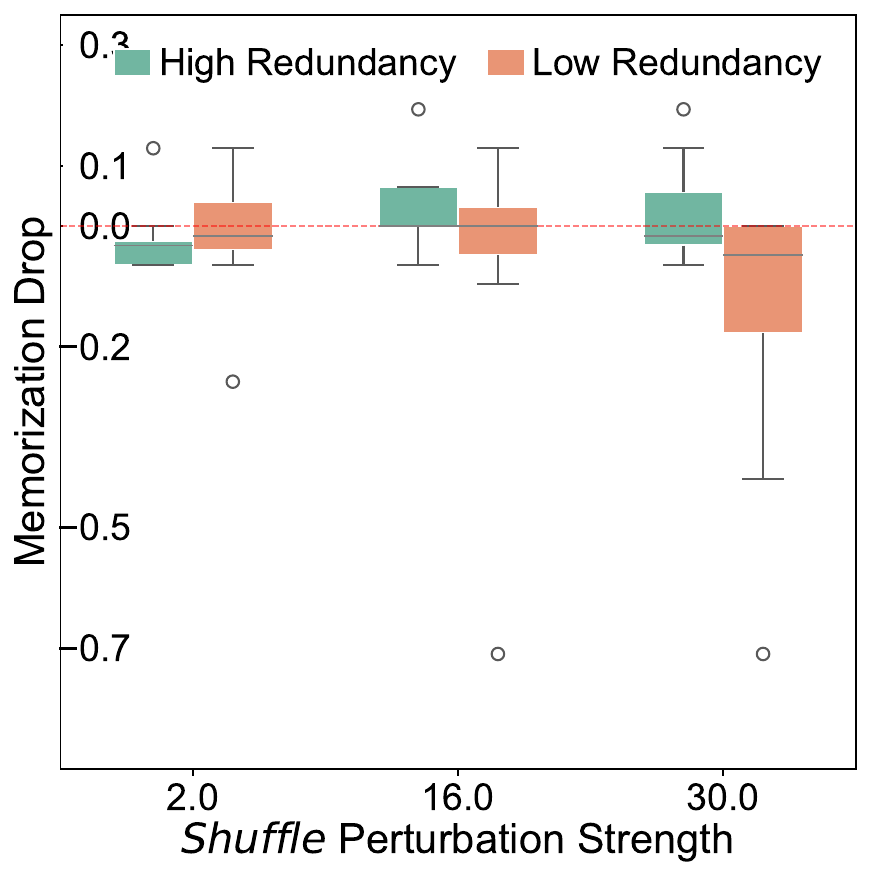}
        \includegraphics[width=0.24\linewidth]{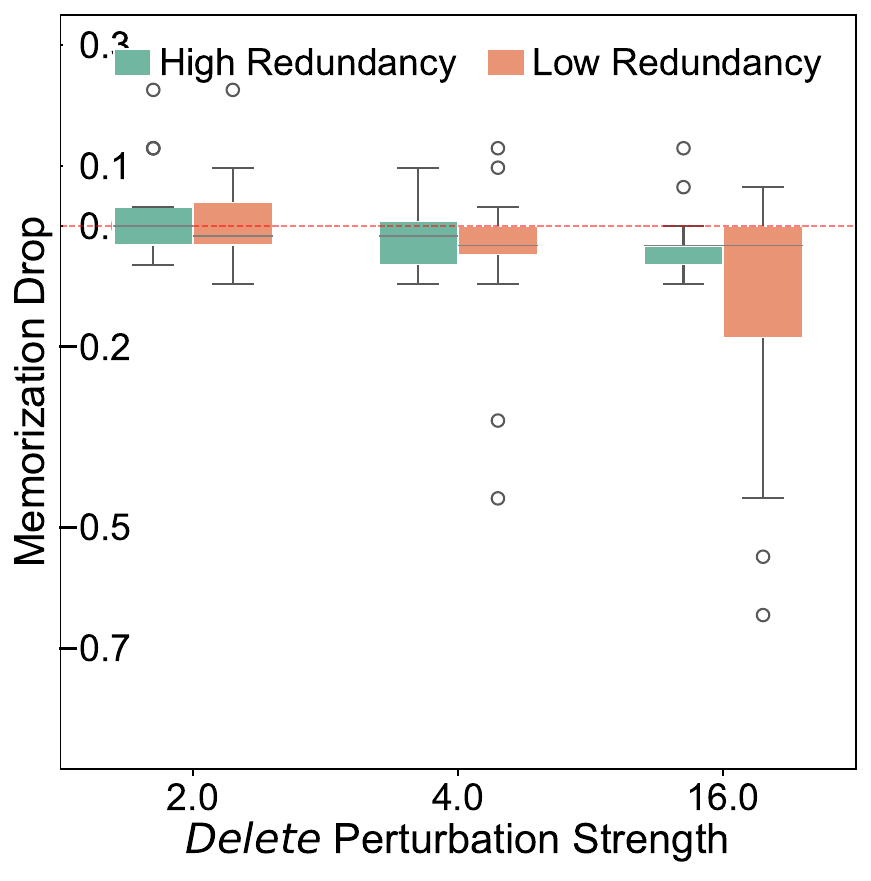}
        \includegraphics[width=0.24\linewidth]{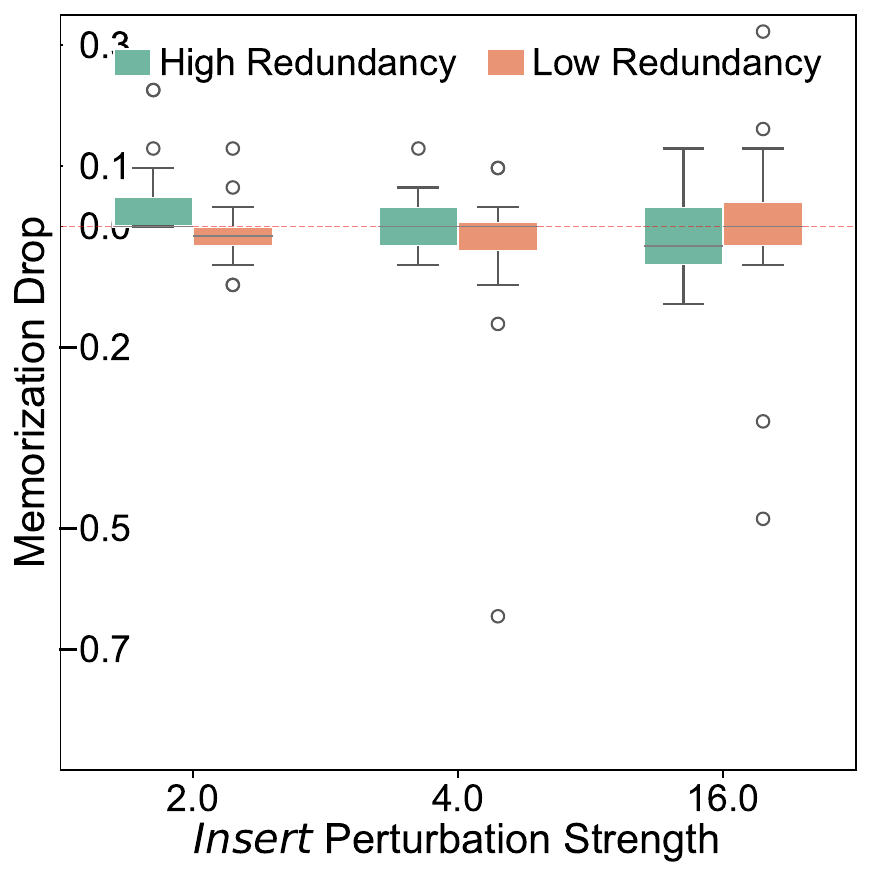}
        \includegraphics[width=0.24\linewidth]{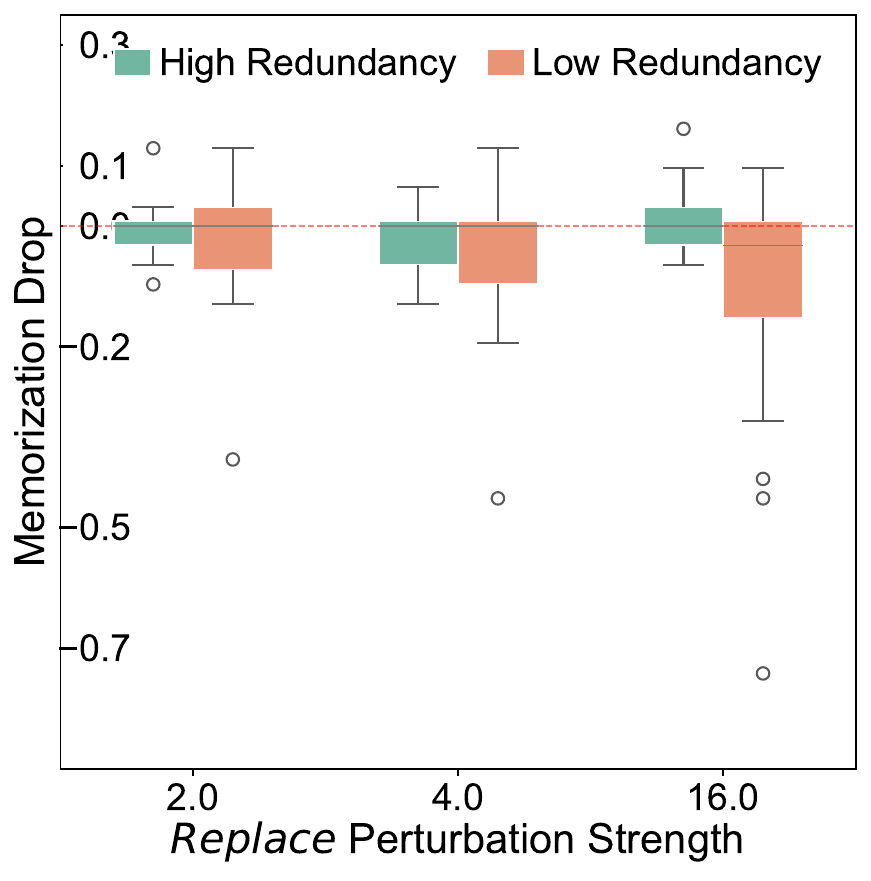}

        \includegraphics[width=0.24\linewidth]{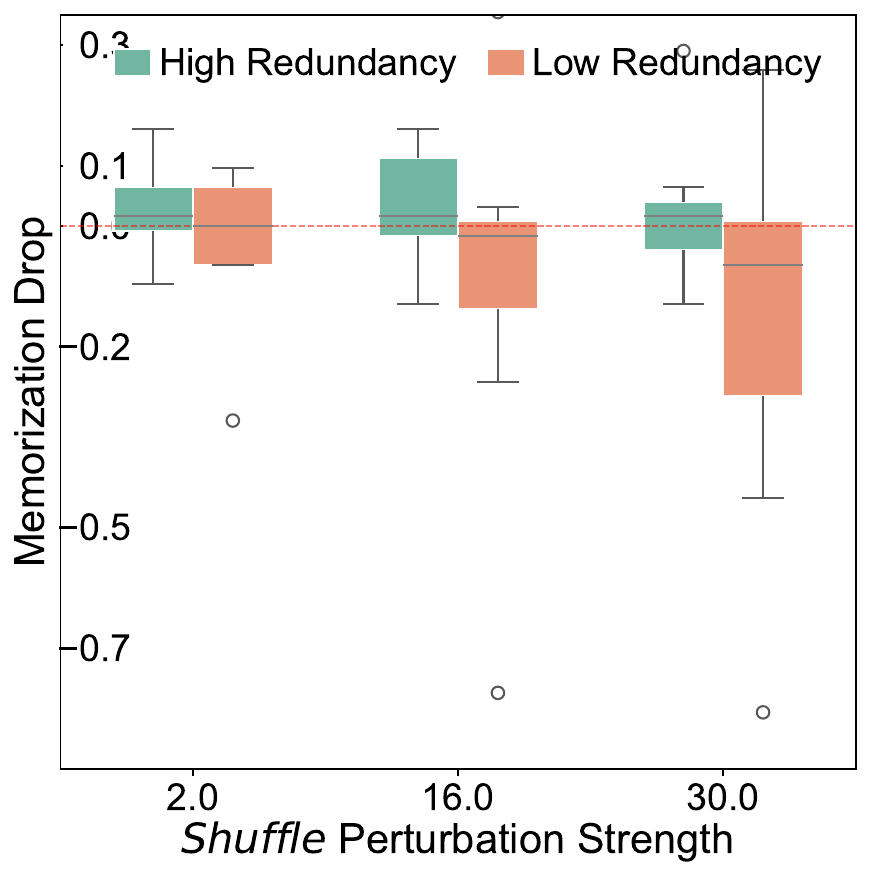}
        \includegraphics[width=0.24\linewidth]{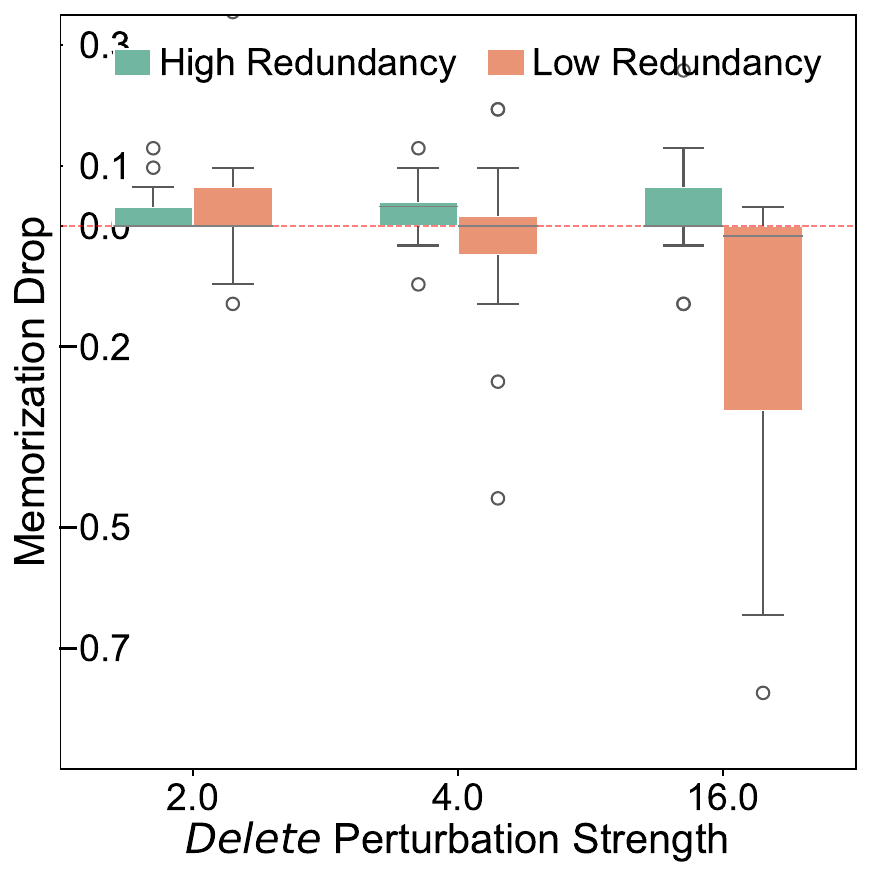}
        \includegraphics[width=0.24\linewidth]{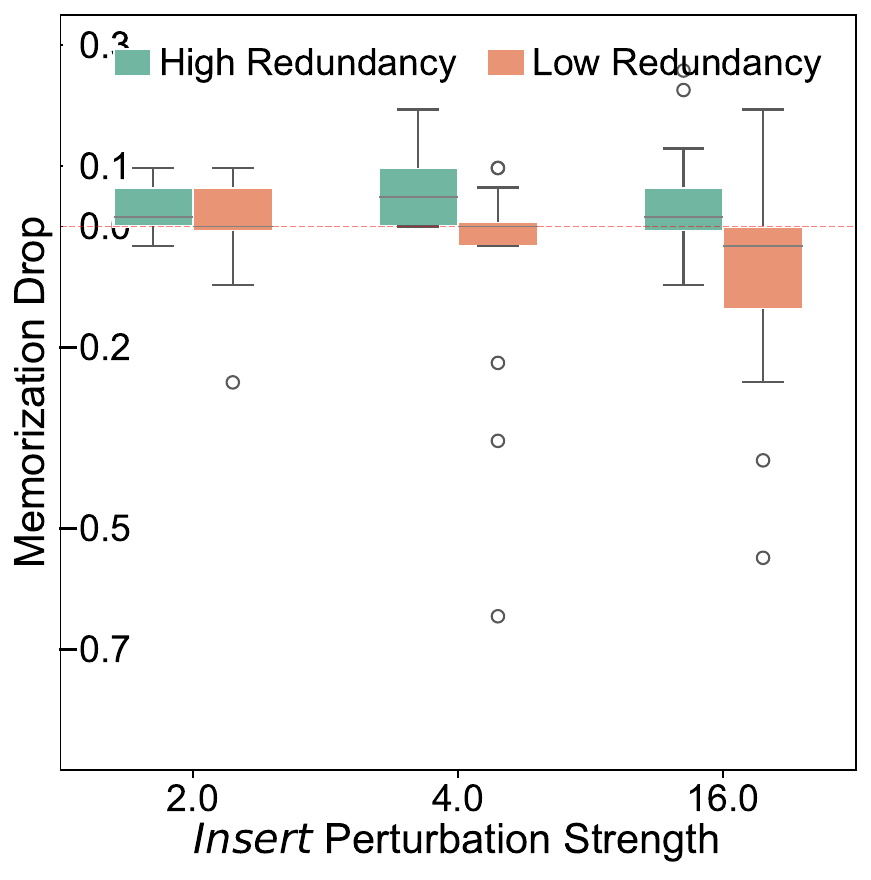}
        \includegraphics[width=0.24\linewidth]{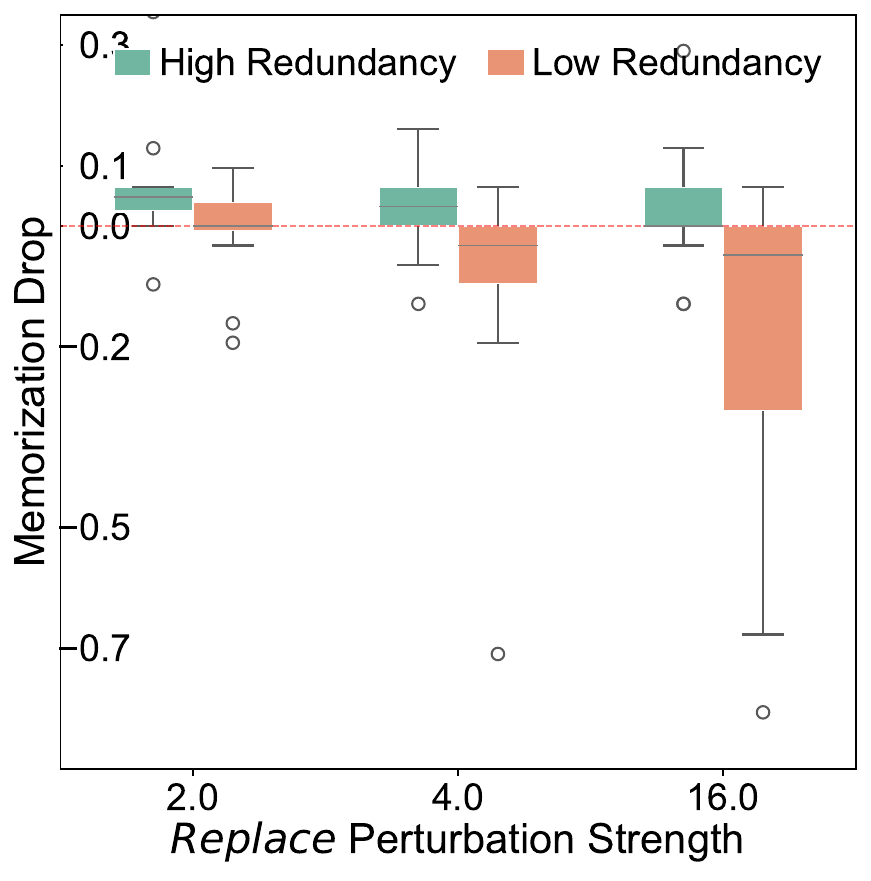}

         \includegraphics[width=0.24\linewidth]{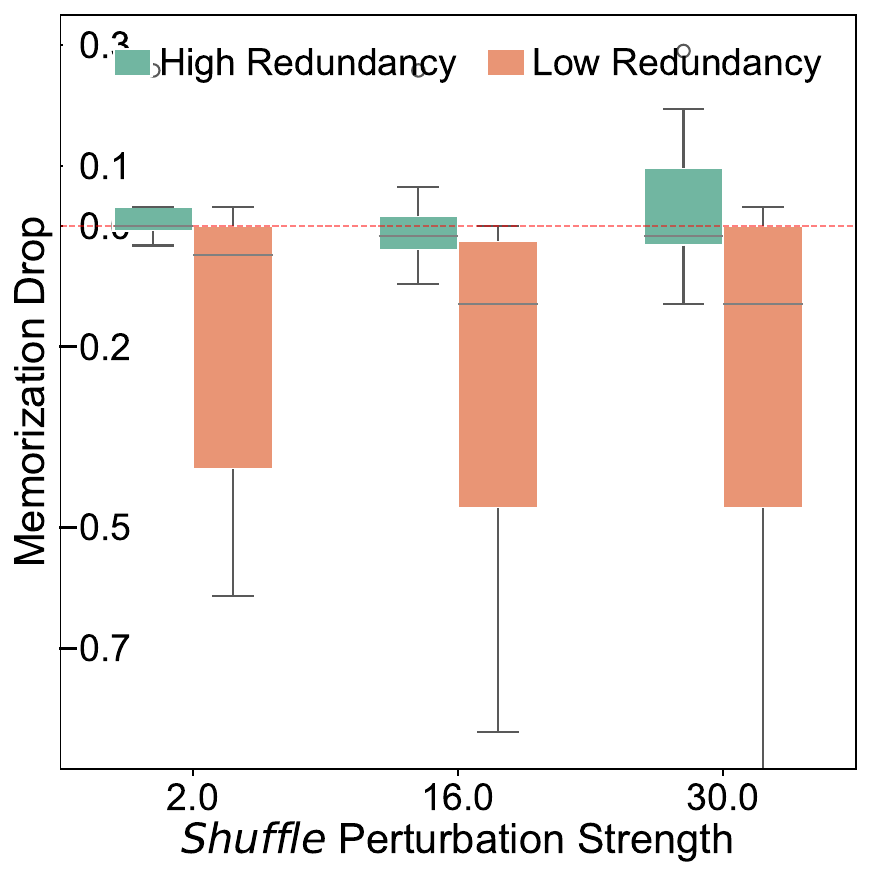}
        \includegraphics[width=0.24\linewidth]{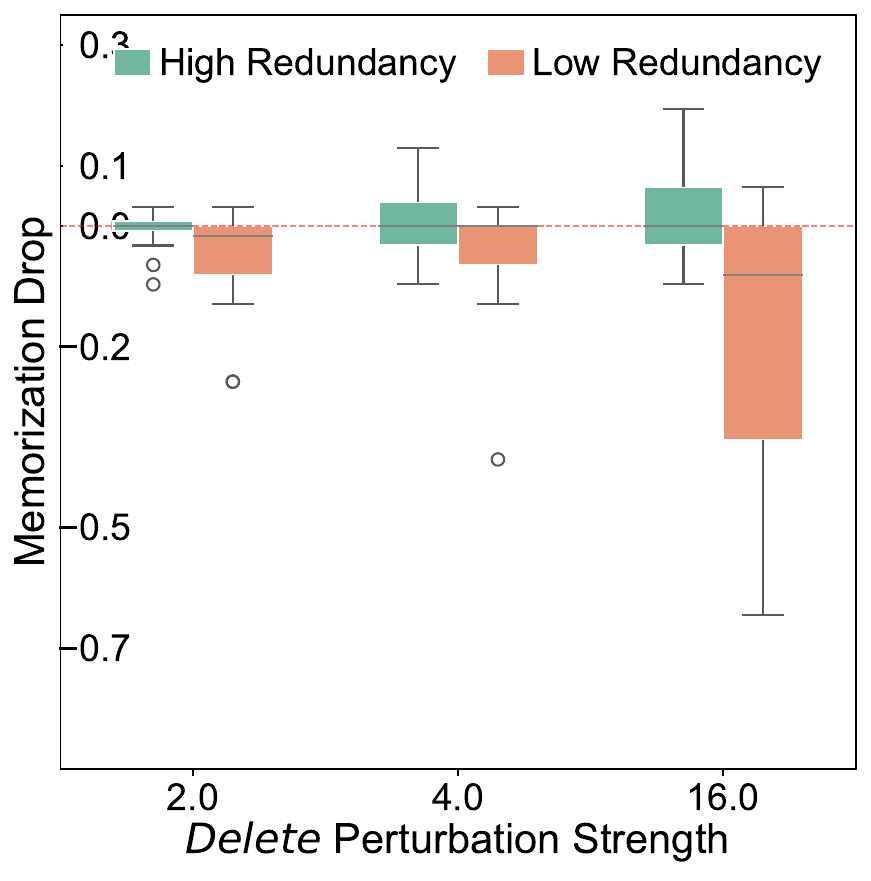}
        \includegraphics[width=0.24\linewidth]{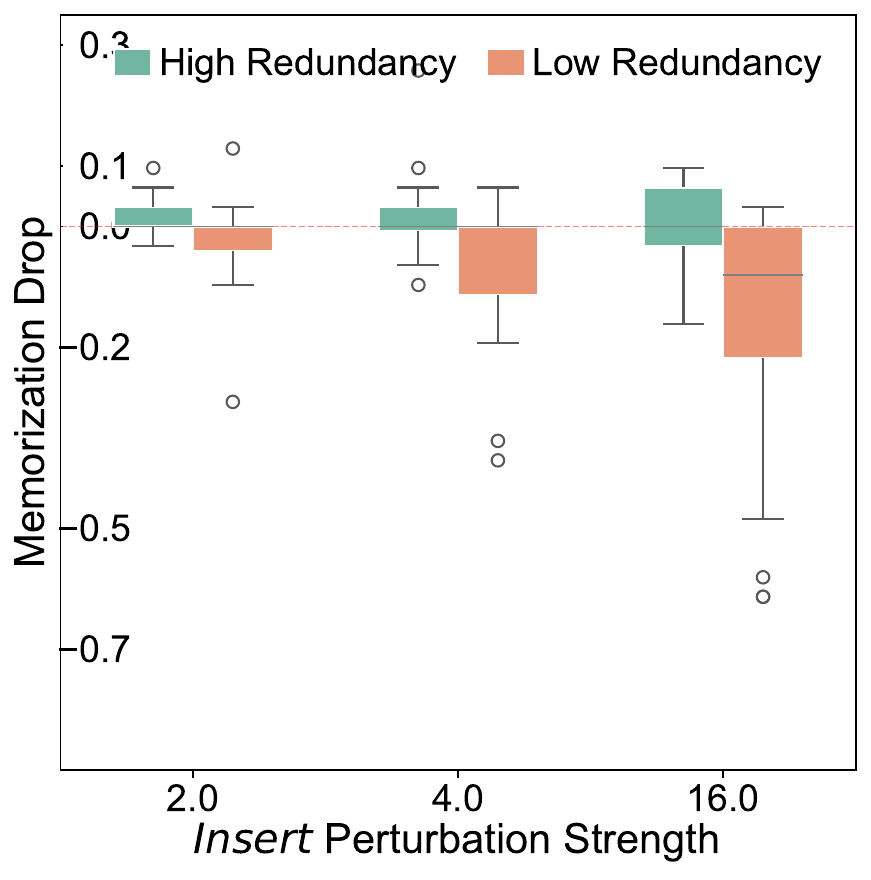}
        \includegraphics[width=0.24\linewidth]{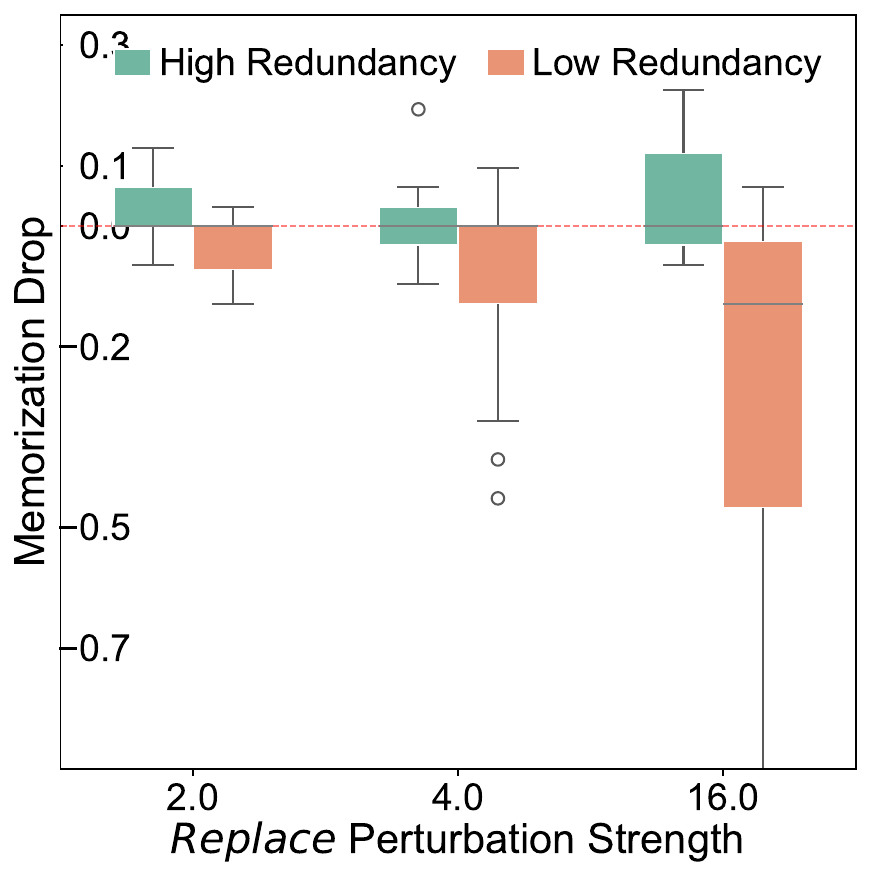}

         \includegraphics[width=0.24\linewidth]{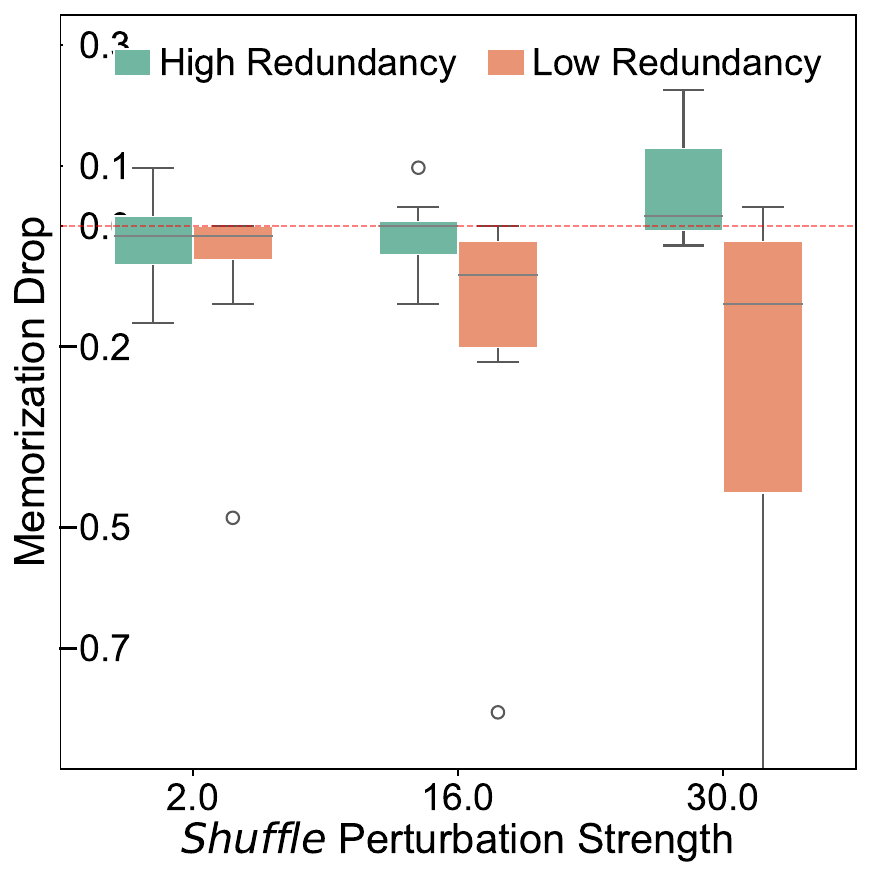}
        \includegraphics[width=0.24\linewidth]{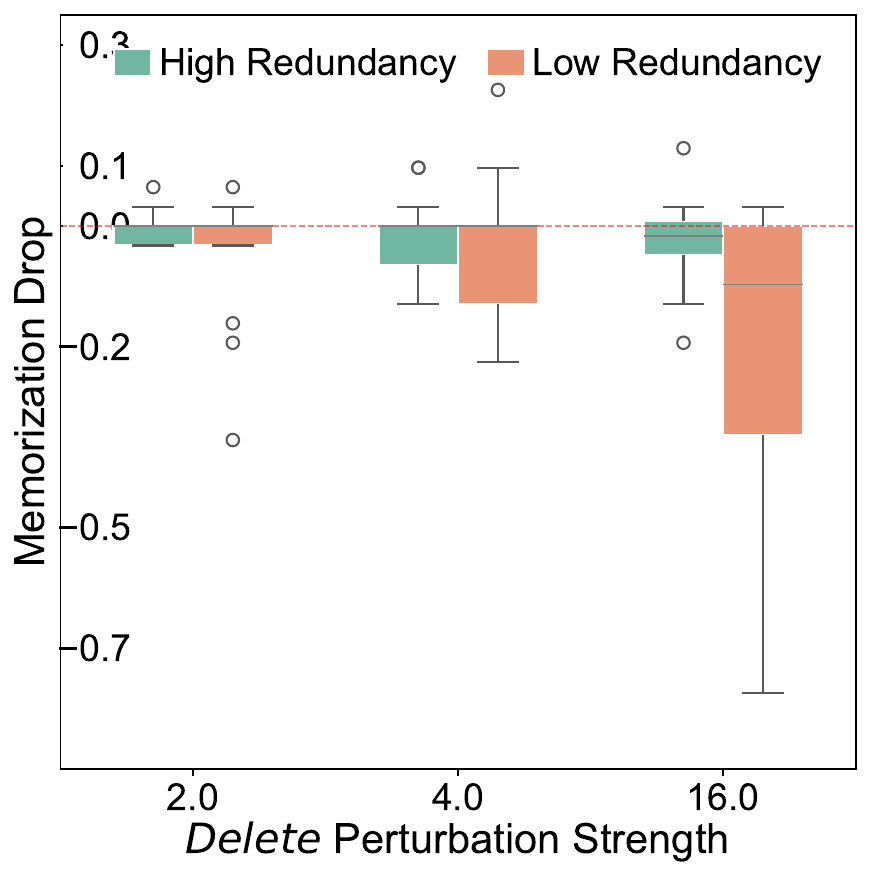}
        \includegraphics[width=0.24\linewidth]{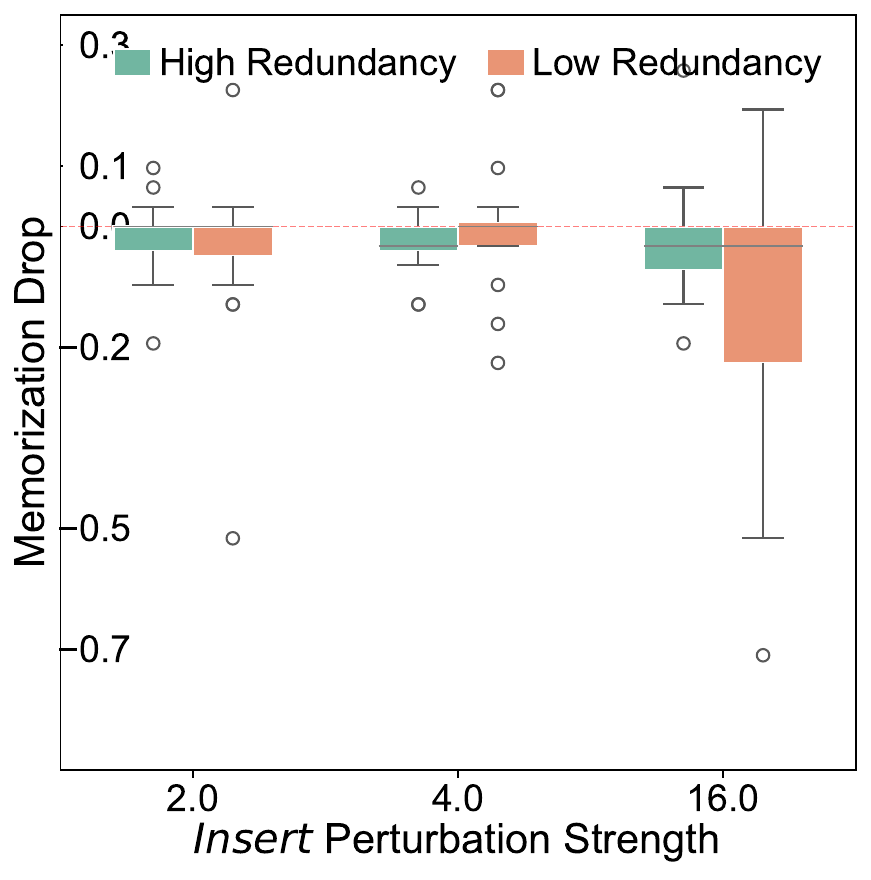}
        \includegraphics[width=0.24\linewidth]{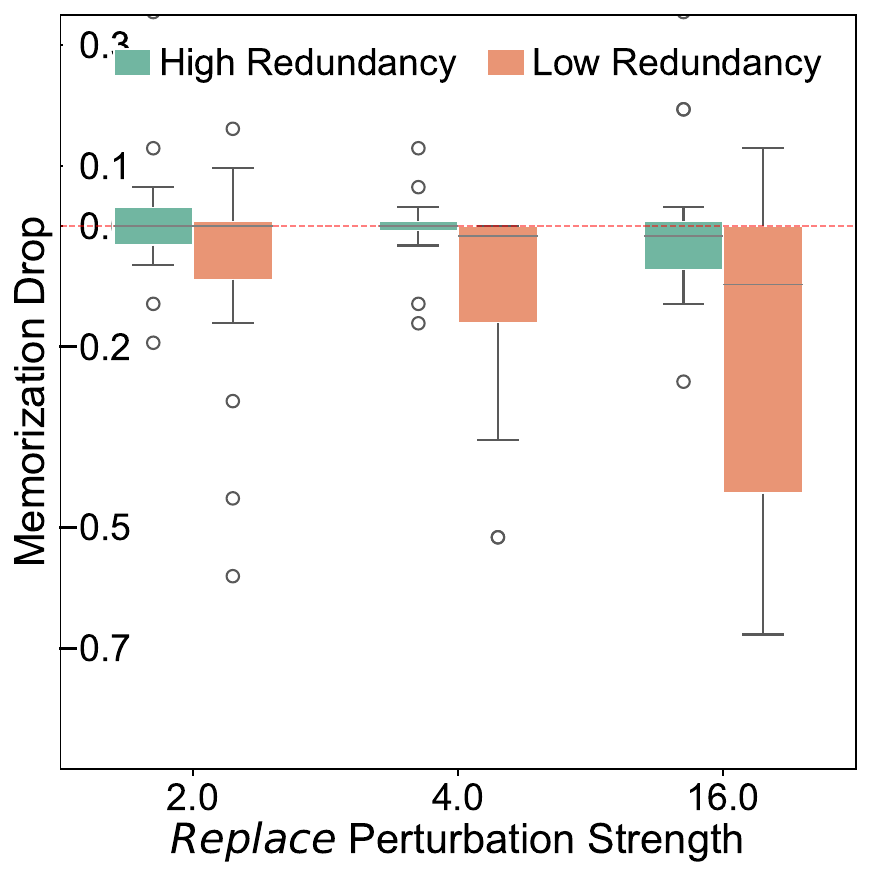}
\caption{Different impacts of perturbation strength on sequence redundancy across Pythia 160M, 410M, 6.9B and 12B (with $t$-test $p < 0.001$).}
\label{fig:low_high_redundancy}
\end{figure}

\section{Beyond Surface-Level Memorization Factors}

\subsection{\textbf{Redundancy and Memorization Vulnerability}} \label{sec:exp_1}

To further analyze the impact of redundancy on memorization,  we partitioned test samples into high- and low-redundancy groups and examined memorization score changes under varying perturbation intensities spanning both small and large scales. Figure \ref{fig:low_high_redundancy} presents our empirical results, where the x-axis represents perturbation strength determined by control factor $r=\{0.1, 0.5, 0.9\}$, i.e., $r*T=\{2, 16, 30\}$, and the y-axis denotes the difference between post- and pre-perturbation memorization scores.
The memorization drop exhibits significant differences between  groups, with lower values for high redundancy samples and higher values for low redundancy samples. This trend holds consistently across all 8 tested models and four perturbation types. Due to space constraints, we present experimental results for four models: 160M, 410M, 6.9B, and 12B. 
Taking Pythia 12B as an example, at strength 16, low redundancy groups show significantly higher memorization changes (i.e., 0.16) compared to high redundancy groups (i.e., 0.08). This disparity amplifies with both increased perturbations and model scale: the memorization drop difference expands from 0.08 to 0.6 as perturbation strength increases from 2 to 30, and from 0.05 to 0.17 as model scale grows from 160M to 12B (strength at 16).
Our findings validate that redundancy correlates with memorization, aligning with established noise resistance principles. This suggests that low-redundancy sequences may be more vulnerable due to their information density, i.e., each token disruption eliminates unique, non-recoverable content. Conversely, high-redundancy sequences withstand modifications through information overlap. 
Statistical analysis confirms these differences are significant across all tested conditions with $p < 0.001$, indicating large effect sizes for redundancy-based vulnerability differences.
% This raises the possibility that redundancy's stronger effect on memorized samples in Figure \ref{fig:Redundancy} could be attributed to a higher proportion of low-redundancy content within memorized samples.

\subsection{\textbf{Low-Redundancy Dominance in Memorized Content}}\label{sec:exp_2}
\begin{figure}[!t]

        \includegraphics[width=0.24\linewidth]{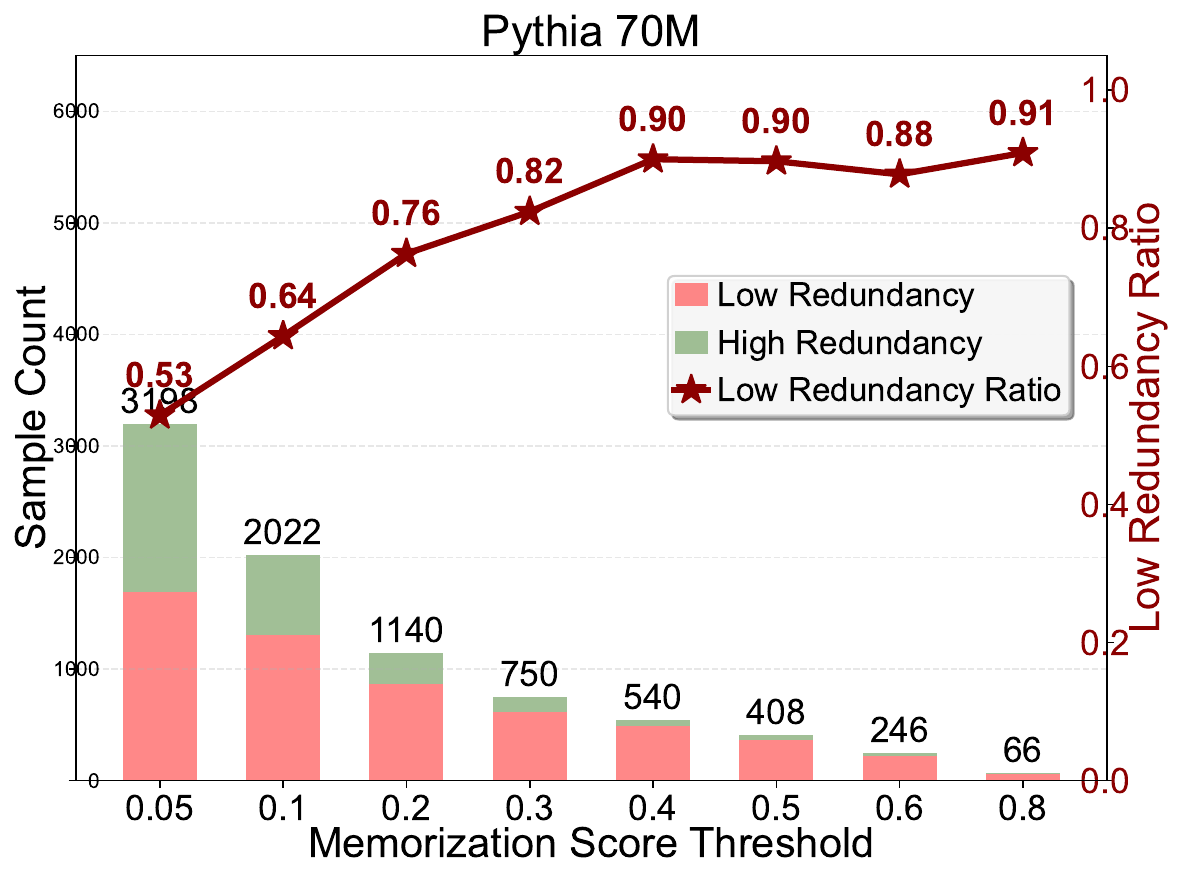}
        \includegraphics[width=0.24\linewidth]{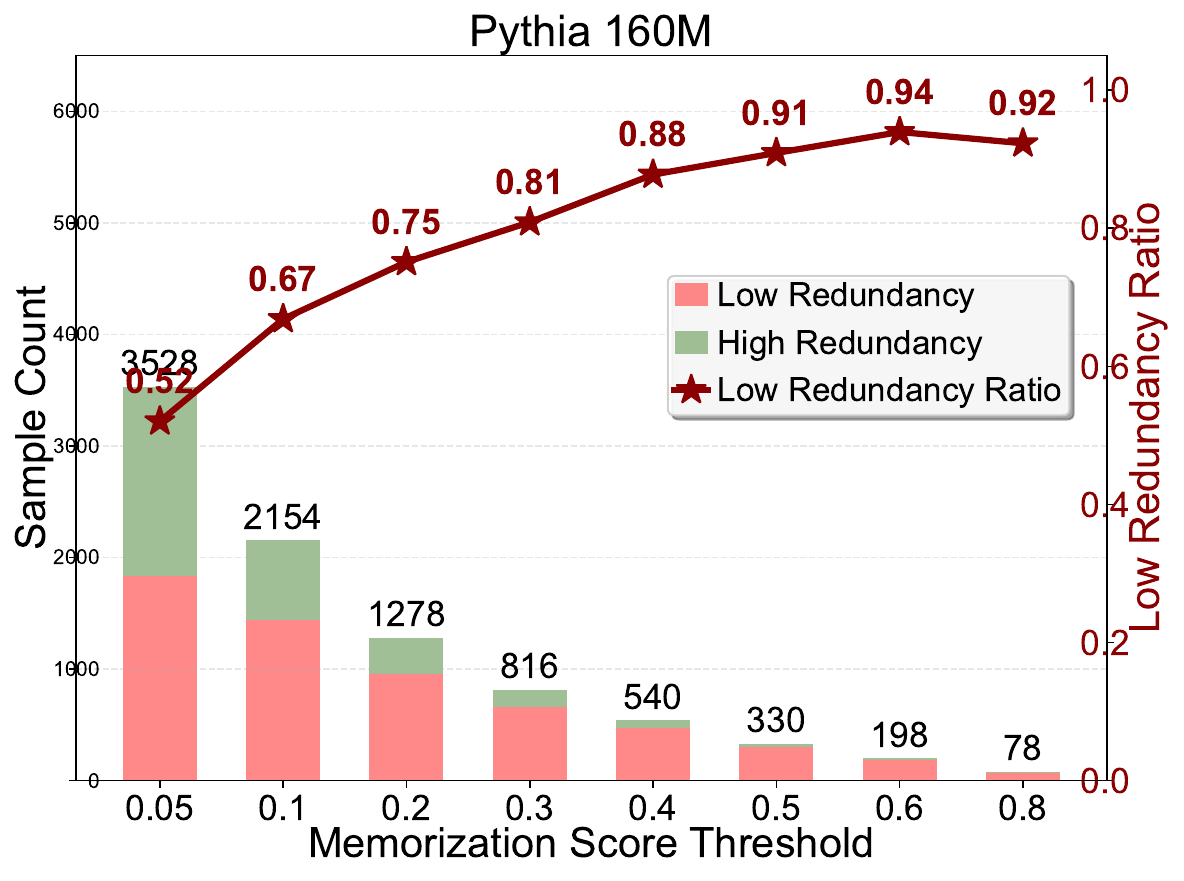}
        \includegraphics[width=0.24\linewidth]{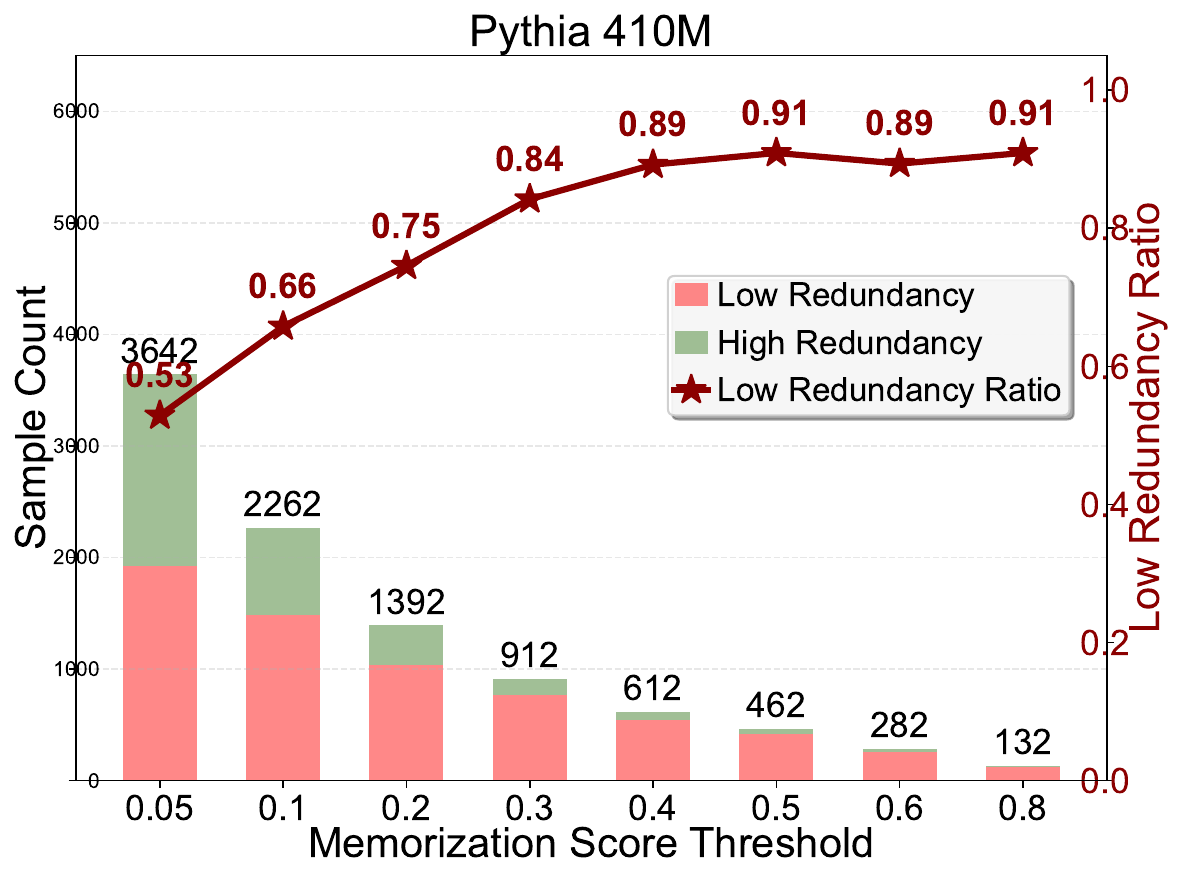}
        \includegraphics[width=0.24\linewidth]{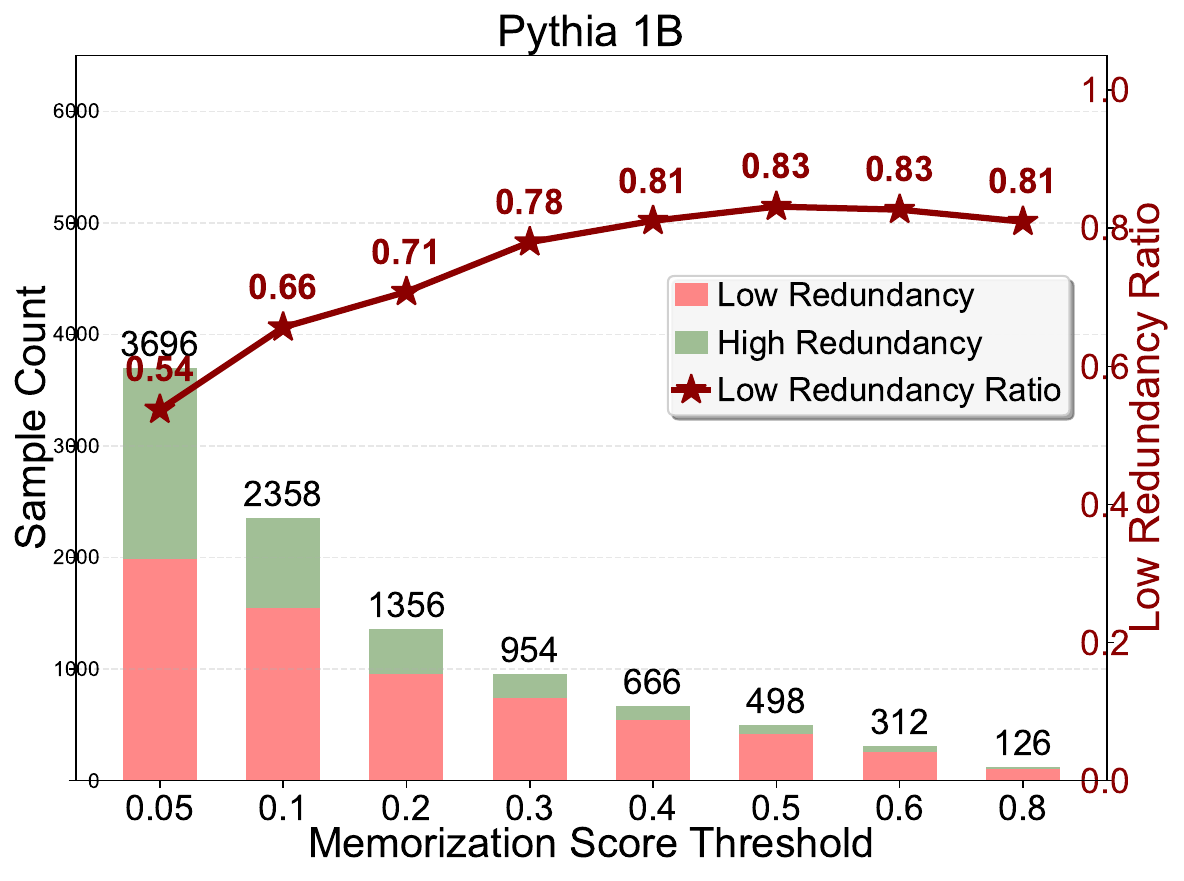}
        \includegraphics[width=0.24\linewidth]{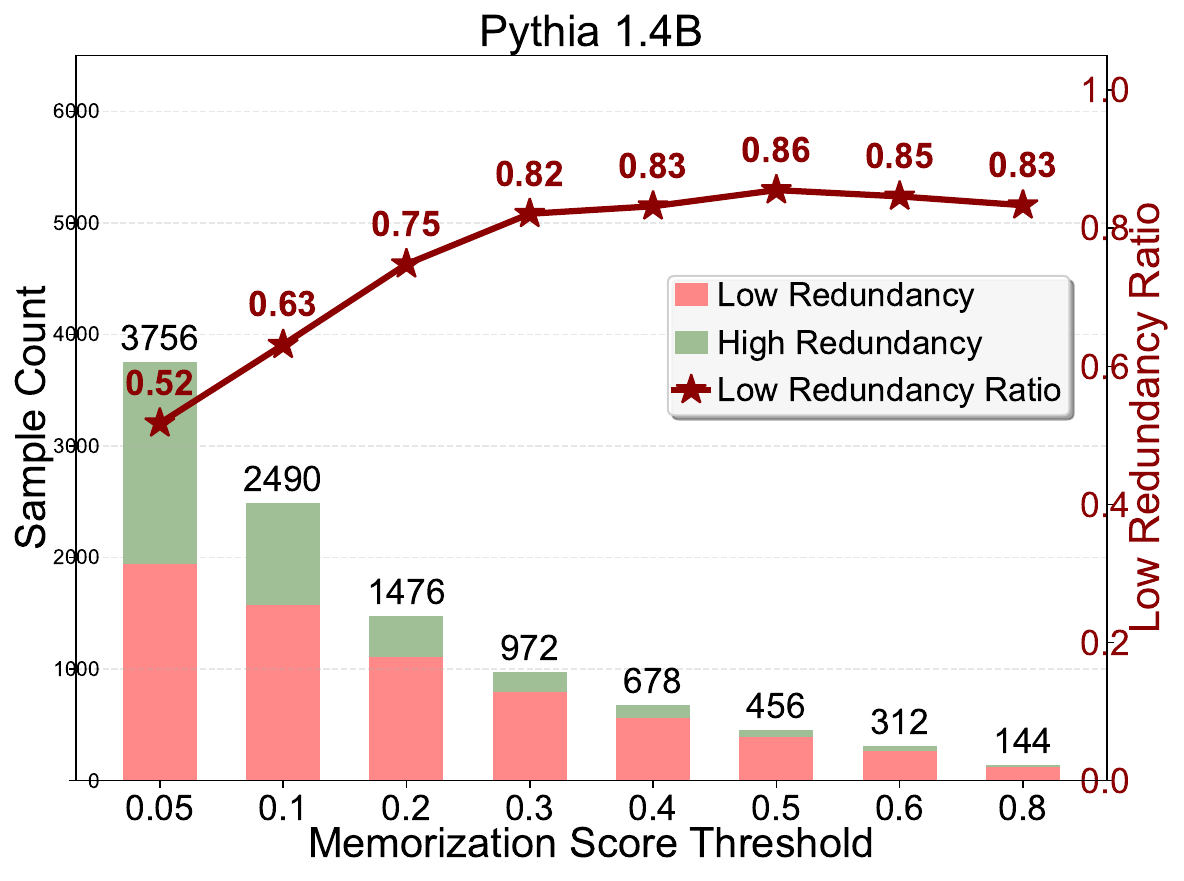}
        \includegraphics[width=0.24\linewidth]{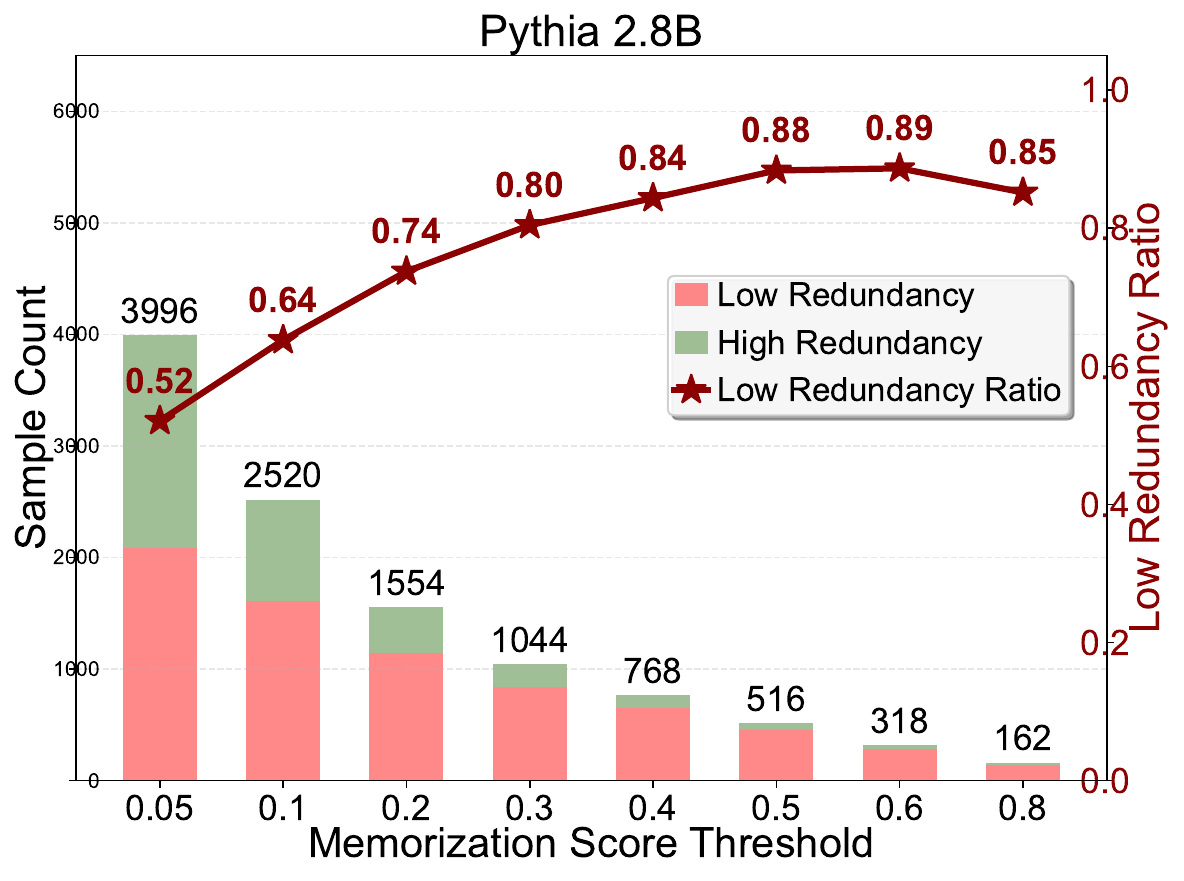}
        \includegraphics[width=0.24\linewidth]{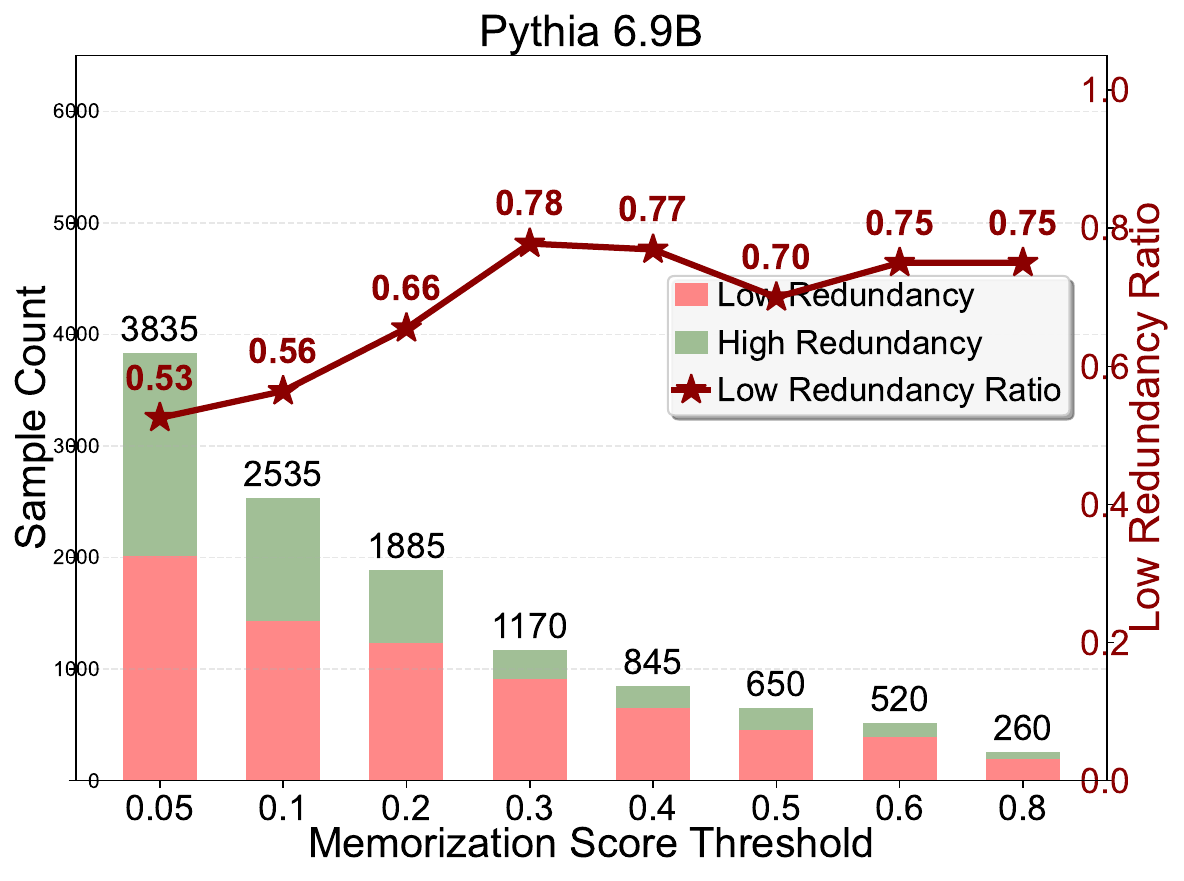}
        \includegraphics[width=0.24\linewidth]{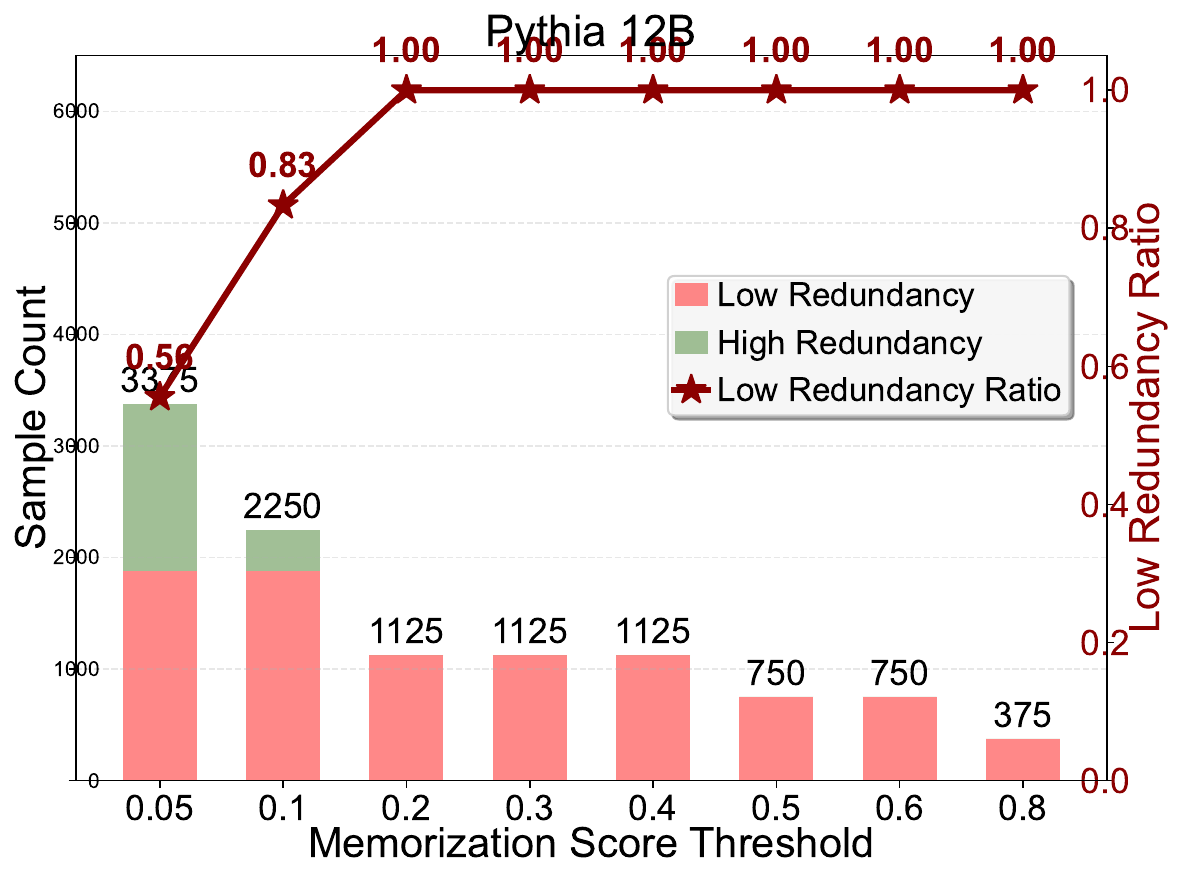}

\caption{Redundancy ratio  at different memorization thresholds ($n=2$).}
\label{fig:sample_count}
\end{figure}

Building on the finding that low-redundancy samples exhibit  higher vulnerability to perturbations, we conjectured that memorized groups predominantly contain low redundancy samples, which would explain the observed fragility patterns and thereby account for their vulnerability to perturbations. 
To test this conjecture, we examined the distribution of high and low redundancy samples within memorized and non-memorized classifications across various $\theta$ thresholds $\{0.05, 0.1, 0.2, 0.3, 0.4, 0.5, 0.6, 0.8\}$. As illustrated in Figure \ref{fig:sample_count}.  The data provides definitive evidence of models' low-redundancy memorization bias: memorized samples contain 79\% low-redundancy versus 21\% high-redundancy samples on average,  consistently observed across all model scales and threshold configurations, confirming  preferential memorization of low-redundancy content (with $\chi^2$  test, $p<0.001$). Note that, consistent trends emerge across different $n=\{1, 2, 3, 5\}$ configurations, we present results for $n=2$ due to space limitations.
% 为什么模型喜欢low redundancy的呢？

% ------------------------

\begin{figure}[!htb]
    \centering

        \includegraphics[width=0.24\columnwidth]{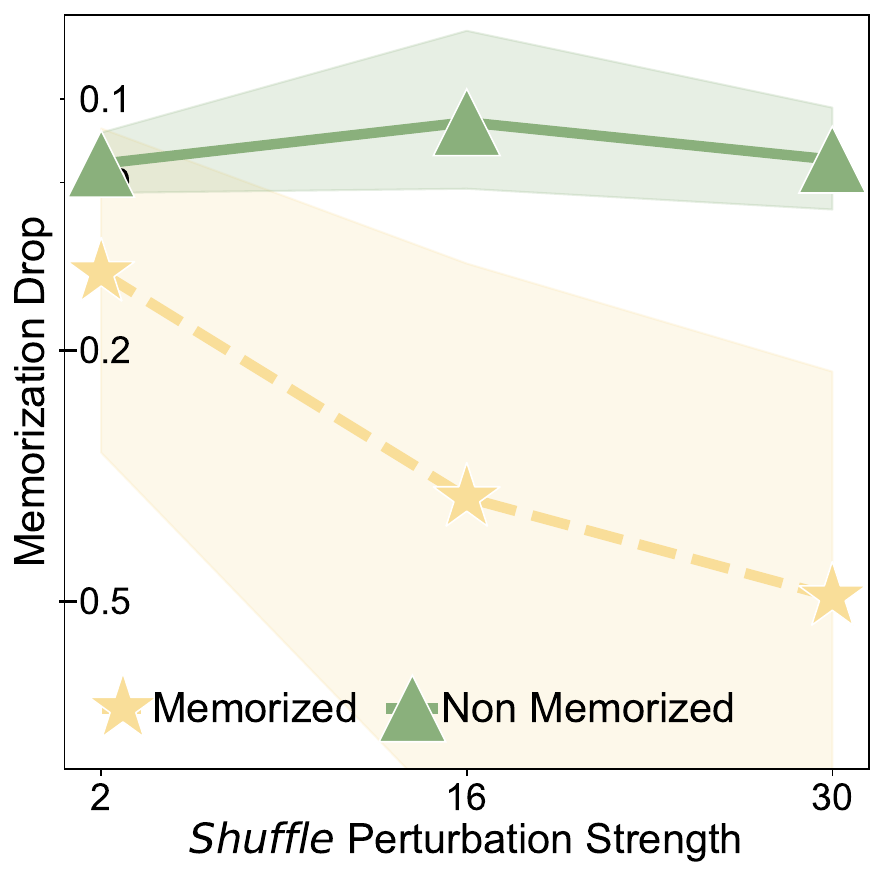}
        \includegraphics[width=0.24\columnwidth]{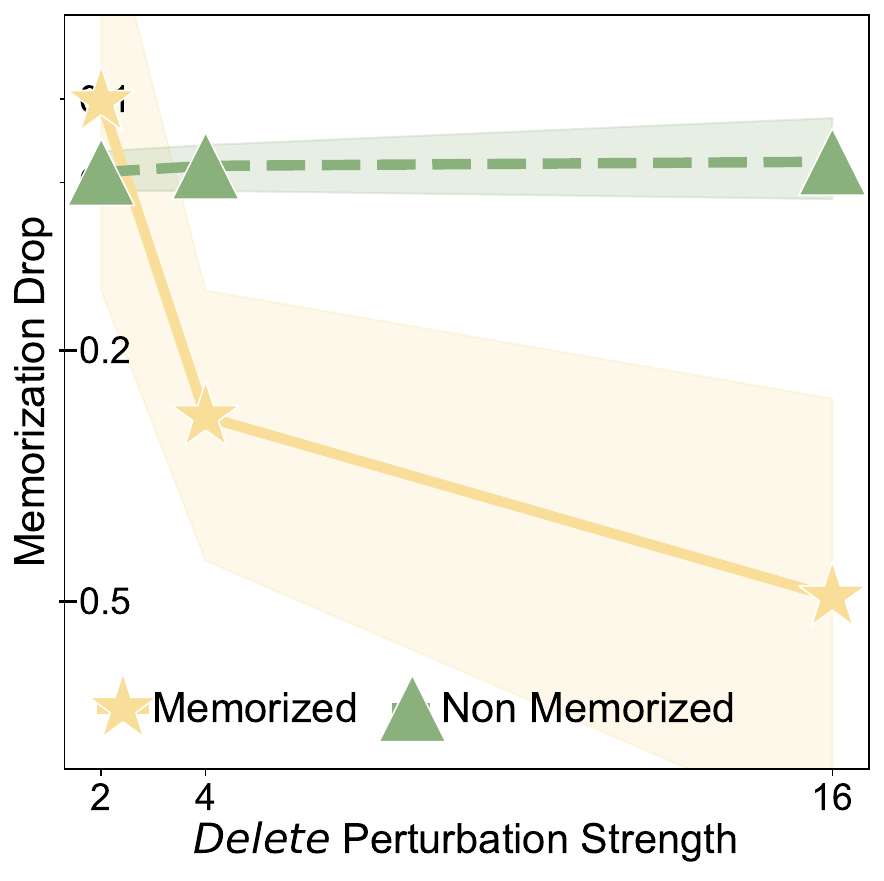}
        \includegraphics[width=0.24\columnwidth]{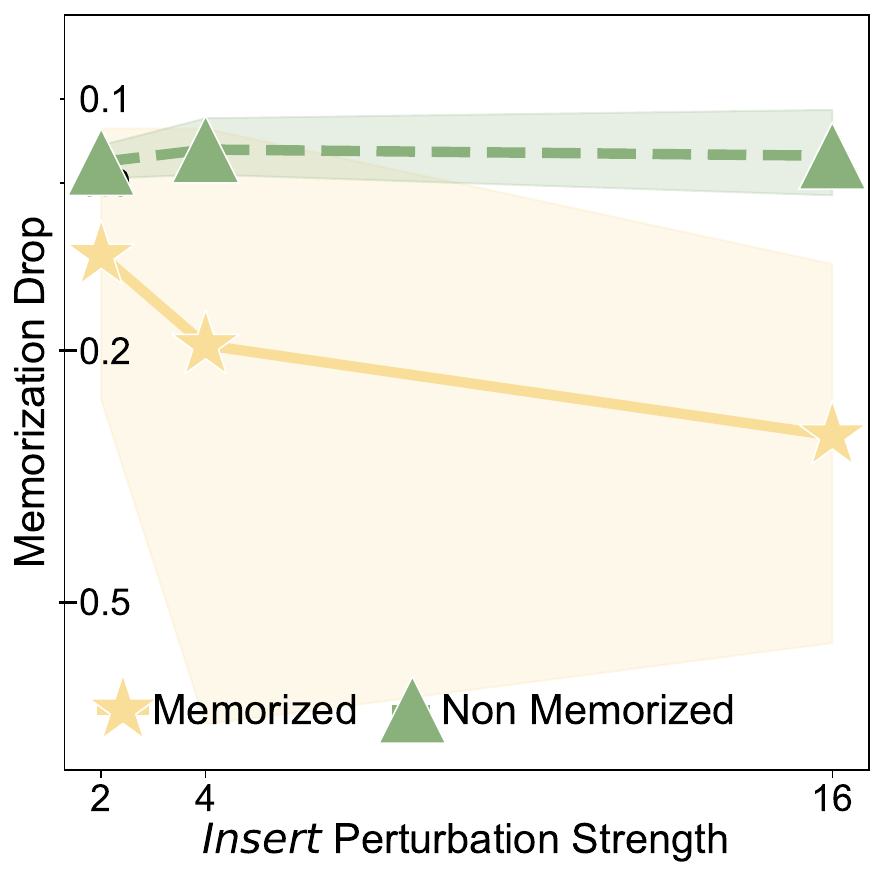}
        \includegraphics[width=0.24\columnwidth]{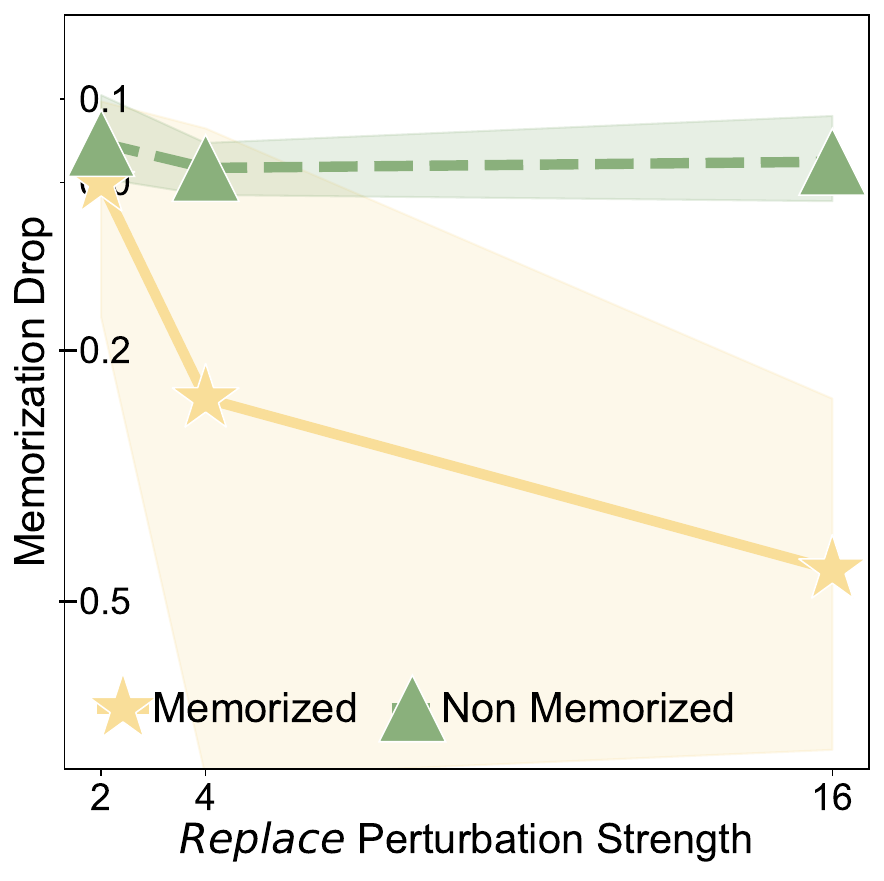}

        \includegraphics[width=0.24\columnwidth]{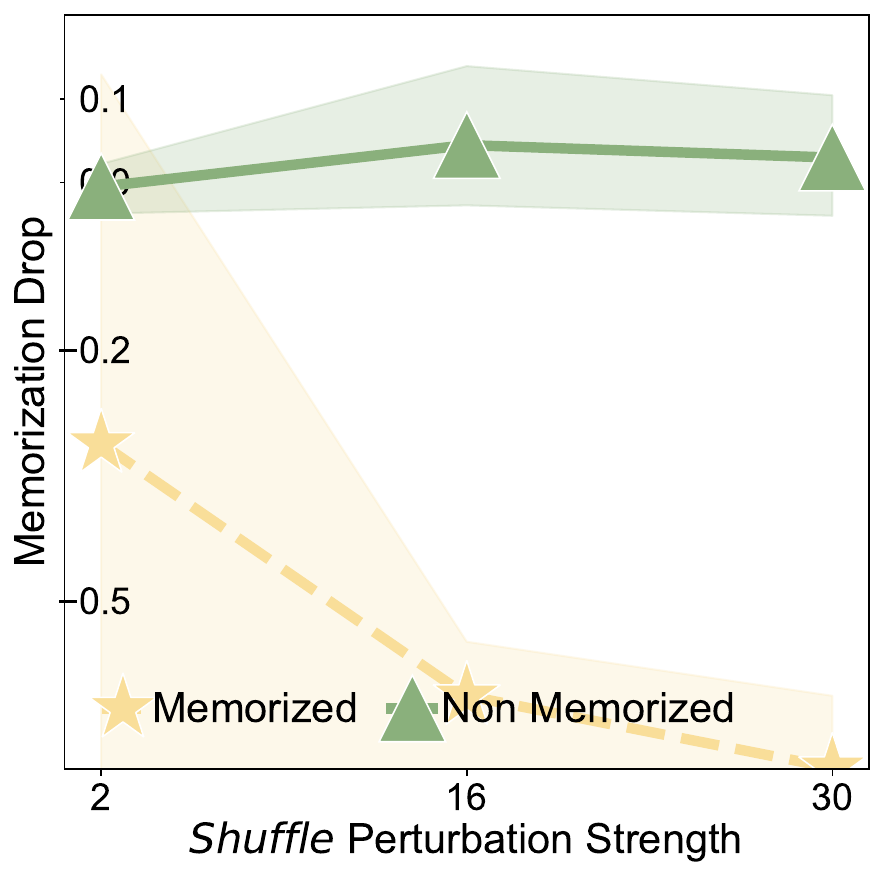}
        \includegraphics[width=0.24\columnwidth]{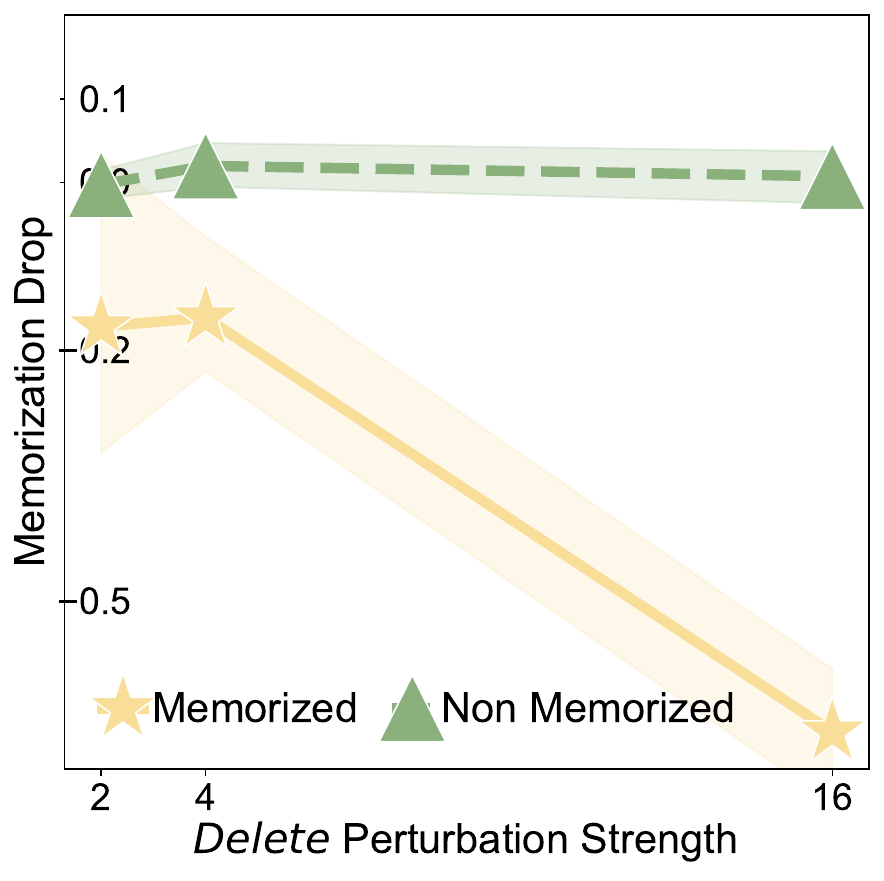}
        \includegraphics[width=0.24\columnwidth]{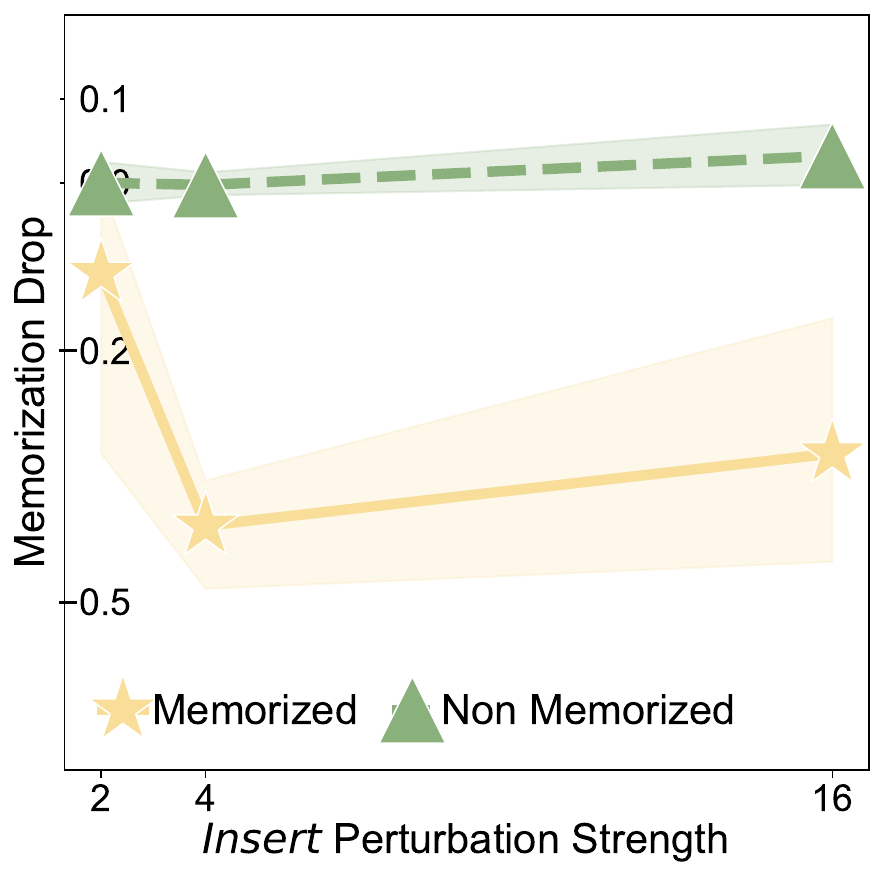}
        \includegraphics[width=0.24\columnwidth]{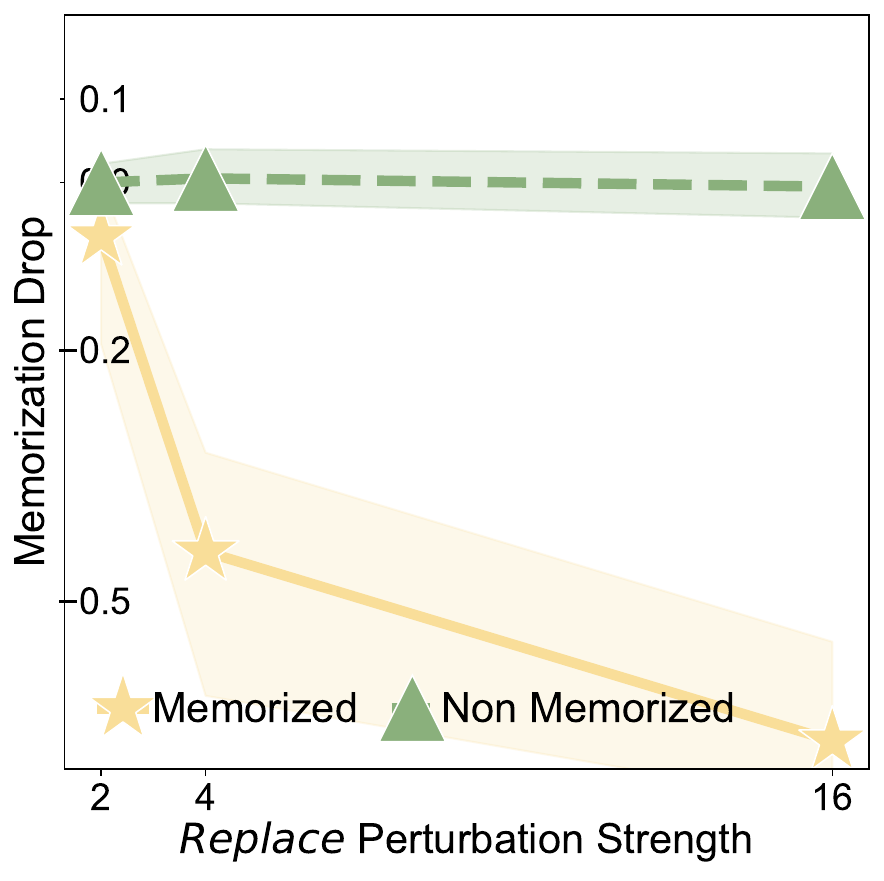}

        \includegraphics[width=0.24\columnwidth]{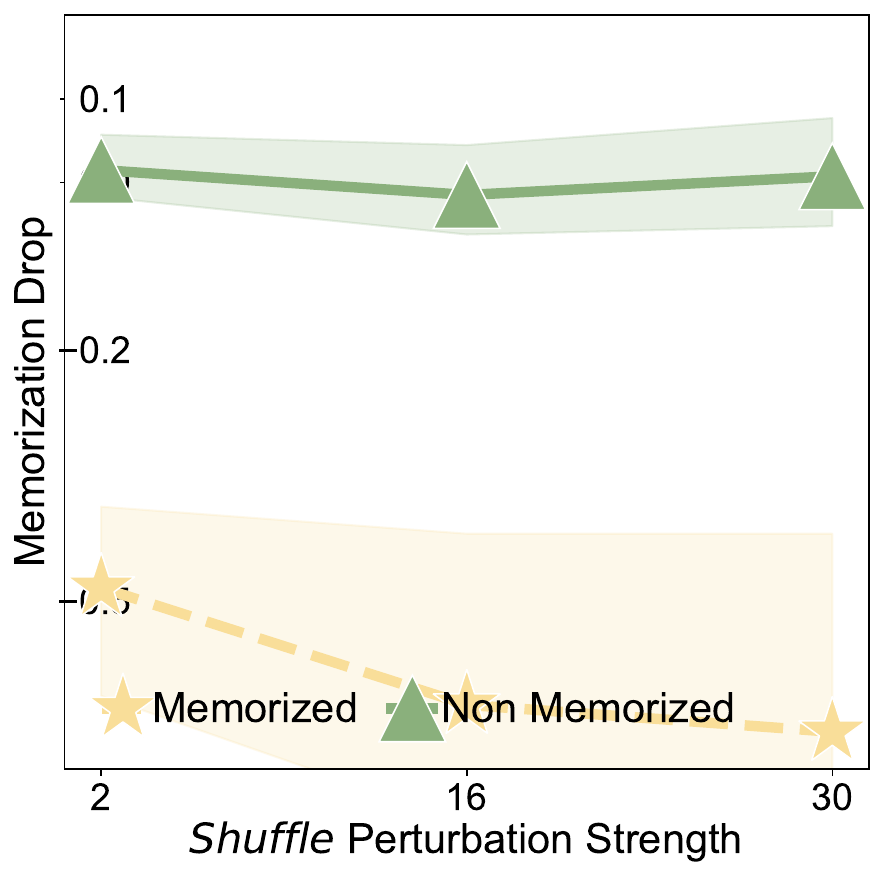}
        \includegraphics[width=0.24\columnwidth]{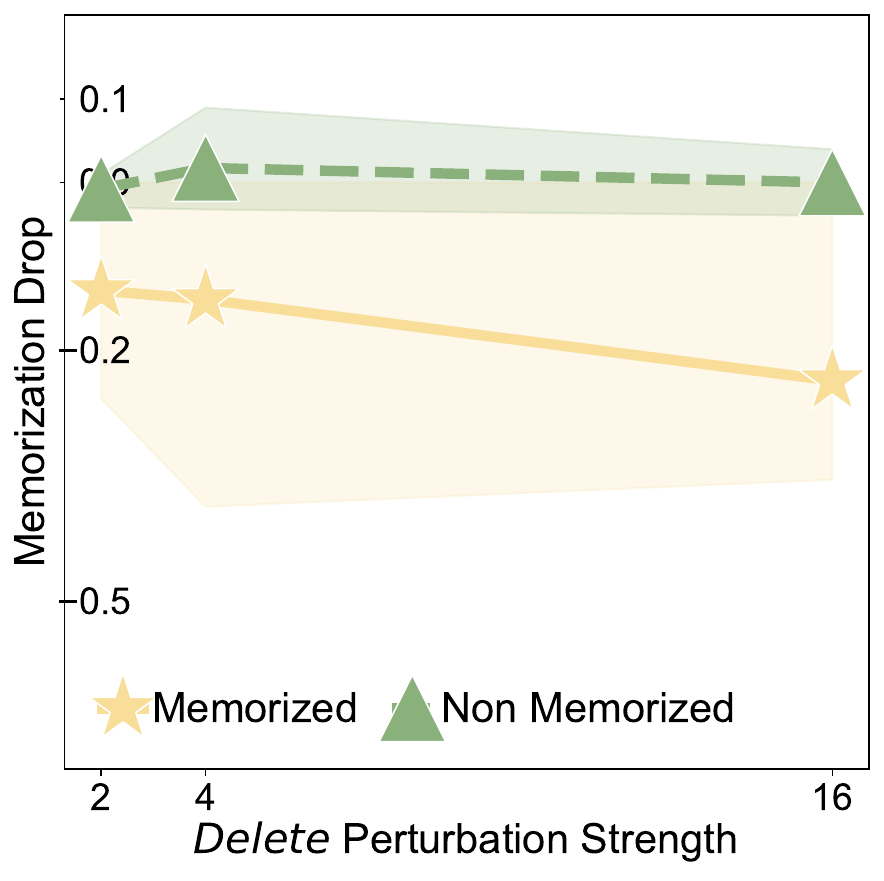}
        \includegraphics[width=0.24\columnwidth]{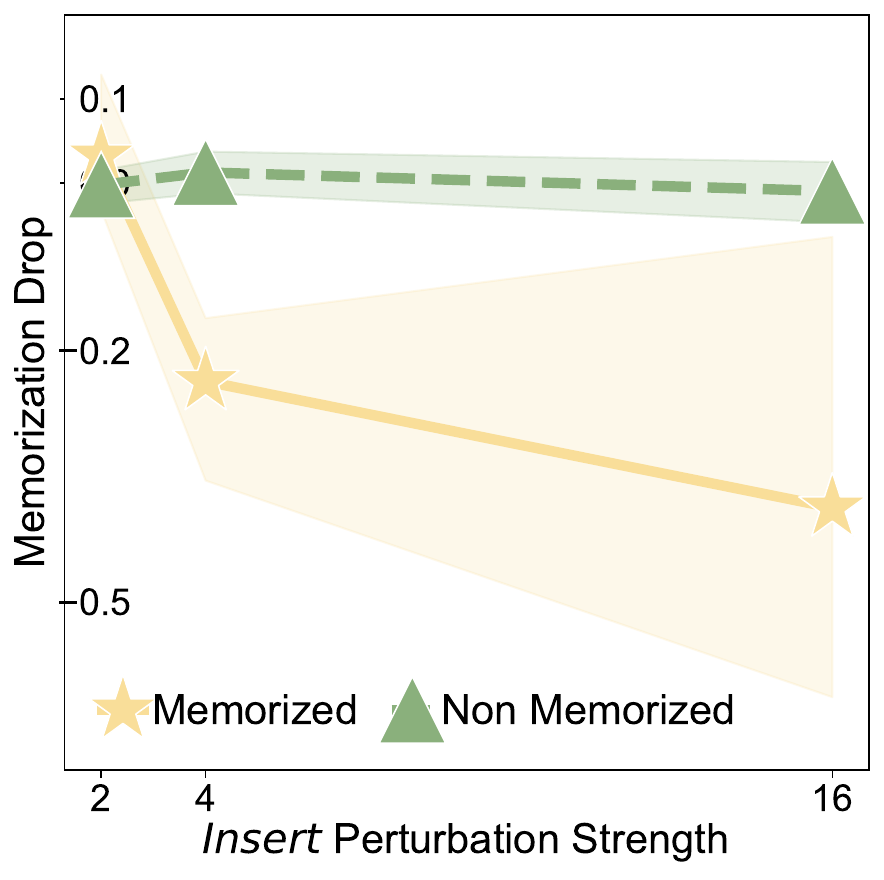}
        \includegraphics[width=0.24\columnwidth]{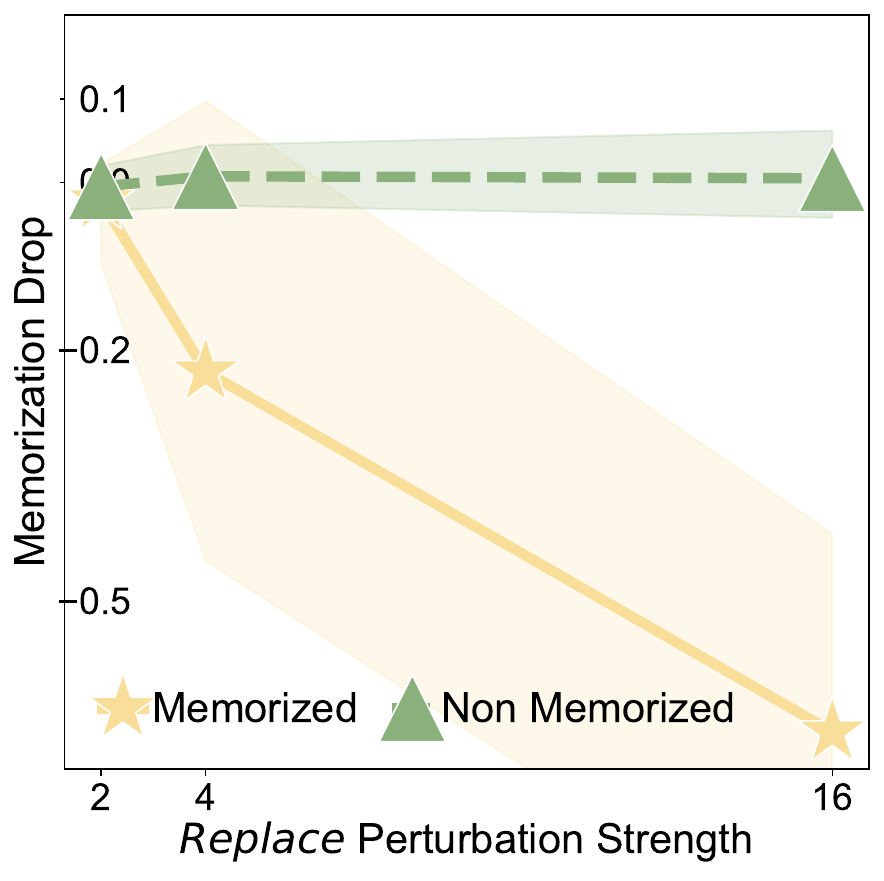}

\caption{Differential Vulnerability of Memorized and Non-Memorized Samples to Perturbations (410M, 1B, 6.9B)}
\label{fig:perturbation_memorization}
\end{figure}

\subsection{Prediction of Low-Redundancy Memorization}\label{sec:exp_3}
The convergent evidence from vulnerability analysis (Section \ref{sec:exp_1} ) and distribution analysis (Section \ref{sec:exp_2}) enables us to formulate a testable prediction: if low-redundancy samples exhibit  higher vulnerability (Section \ref{sec:exp_1}) and memorized samples contain 79\% low-redundancy content (Section \ref{sec:exp_2}), then memorized samples as a whole should demonstrate higher perturbation sensitivity compared to non-memorized samples.
We tested this prediction by directly comparing memorization drops between memorized and non-memorized groups under identical perturbations. Using $\theta=0.5$ to partition samples, we apply identical perturbations and measure memorization drops. Figure \ref{fig:perturbation_memorization} confirms this prediction, where memorized samples show sharp, continuous decline while non-memorized samples remain stable near 0.0. At $r*T=30$ replacement perturbations on Pythia 12B, the memorization drops are 0.6 versus 0.01 respectively, representing a substantial difference in vulnerability. {These findings indicate that redundancy is a significant factor in influencing LLM memorization behavior, complementing existing surface-level characteristics and providing mechanistic insight into memorization  patterns.} The trend is consistent across other thresholds, and only the results of $\theta=0.5$ is presented due to space limitation.
% 为什么模型喜欢记忆低冗余度的样本啊？？？？？

\subsection{\textbf{Evolution of Information Redundancy Preferences}}

\begin{figure}[!htb]
    \centering
        \includegraphics[width=0.48\linewidth]{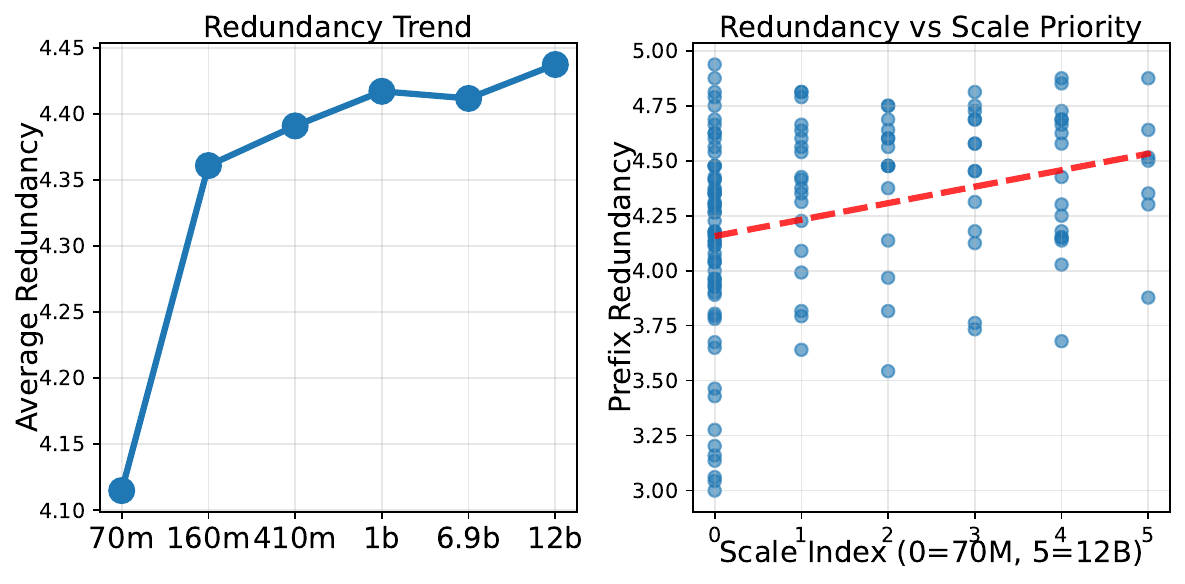}
        \includegraphics[width=0.48\linewidth]{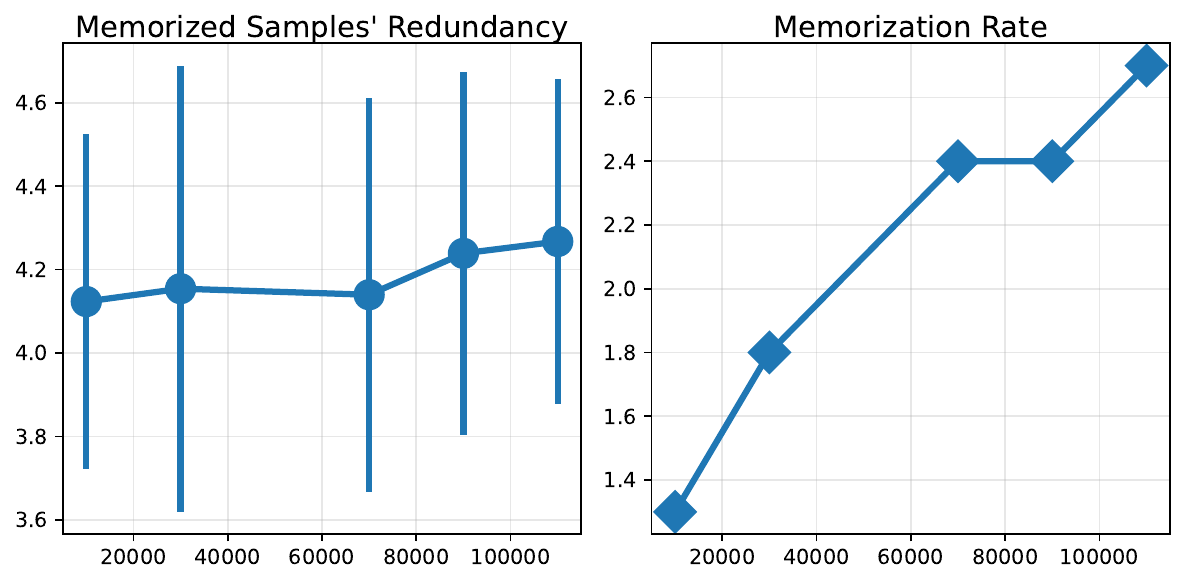}
    \vspace{-3mm}
    \caption{Evolution of Memorization Preferences with Model Scale and Training.}
    \label{fig:information_density_analysis}
\end{figure}

While our analysis demonstrates  low-redundancy dominance in memorized samples, a critical question remains: how does this preference evolve as models develop greater capacity and experience? We investigated memorization patterns across both model scale (including Pythia 70M to 12B parameters) and training progression (including 20K to 120K trained steps) to understand the dynamics of preferences to redundancy.
Figure \ref{fig:information_density_analysis} reveals  evolution patterns. As model scale increases (Figure \ref{fig:information_density_analysis}, first subfigure), average redundancy rises from 4.15 (70M) to 4.45 (12B), indicating that larger models increasingly favor high redundancy compared to smaller models. The redundancy analysis shows that while individual sample redundancy varies widely (scatter in middle-left panel), the overall trend is that samples with lower redundancy are first memorized as scale and training increases.

Temporal evolution analysis (Figure \ref{fig:information_density_analysis}, right panels) demonstrates complementary dynamics. Across training steps from 20K to 100K, memorized samples' redundancy remains relatively stable around 4.0$\sim$4.2, while memorization rates show  growth from 1.4\% to 2.6\%. This suggests that the low-redundancy preference emerges early and remains consistent throughout training.
These findings indicate that as model scale increases and training progresses, models preferentially memorize low redundancy samples first and then gradually memorize more higher redundancy samples.

\vspace{-2mm}
\section{Conclusion}
This work established information redundancy as a  factor underlying memorization behavior in large language models, extending beyond traditional surface-level characteristics. Through  analysis across model scales and training dynamics, we discovered that established memorization factors exhibit differential effects on memorized versus non-memorized samples, revealing that $~$79\% of memorized samples are low-redundancy with significant higher vulnerability than high-redundancy content. Our unified perturbation quantification framework demonstrates that models  prefer low-redundancy content, with this preference weakening  as models scale and mature. These findings provided both theoretical insights into memorization mechanisms and practical guidance for redundancy-guided data preprocessing, enabling more effective privacy risk mitigation and bias reduction in large language models.

% \subsection{RQ3.2:}
% \begin{figure}
%     \centering
%     \begin{subfigure}[b]{0.49\linewidth}
%         \centering
%         \includegraphics[width=\linewidth]{figures/checkpoint_evolution_analysis_160m.pdf}
%         \caption{xxx}
%     \end{subfigure}
%     \begin{subfigure}[b]{0.49\linewidth}
%         \centering
%         \includegraphics[width=\linewidth]{figures/checkpoint_evolution_analysis_410m.pdf}
%         \caption{}
%     \end{subfigure}
    
%     \begin{subfigure}[b]{0.49\linewidth}
%         \centering
%         \includegraphics[width=\linewidth]{figures/checkpoint_evolution_analysis_1.4b.pdf}
%         \caption{xxx}
%     \end{subfigure}
%     \begin{subfigure}[b]{0.49\linewidth}
%         \centering
%         \includegraphics[width=\linewidth]{figures/checkpoint_evolution_analysis_2.8b.pdf}
%         \caption{}
%     \end{subfigure}
    
%     \begin{subfigure}[b]{0.49\linewidth}
%         \centering
%         \includegraphics[width=\linewidth]{figures/checkpoint_evolution_analysis_6.9.pdf}
%         \caption{xxx}
%     \end{subfigure}
%     \begin{subfigure}[b]{0.49\linewidth}
%         \centering
%         \includegraphics[width=\linewidth]{figures/checkpoint_evolution_analysis_12b.pdf}
%         \caption{}
%     \end{subfigure}
%     \caption{}
%     \label{}
% \end{figure}

\section*{Acknowledgments}

This work was partially supported by the Hefei College Talent Research Fund Project (No. 24RC20, 21-22RC13), the National Natural Science Foundation of China Youth Fund Project (No. 62403149), and the Natural Science Foundation of the Anhui Higher Education Institutions of China (No. 2022AH051779).

% \appendix

% \section{Sample Appendix Section}

 % \bibliographystyle{elsarticle-num} 
  \bibliographystyle{elsarticle-num-names} 
 \bibliography{cas-refs}





% \end{thebibliography}
\end{document}